\documentclass[journal]{IEEEtran}
\usepackage{amssymb, amsmath, amsthm, graphicx, enumerate}
\usepackage{setspace}
\usepackage{algorithm}
\usepackage{algpseudocode}
\usepackage{array}
\usepackage{subfigure}
\usepackage{float}
\usepackage{url}
\usepackage{multirow}
\usepackage{stfloats}
\usepackage[none]{hyphenat}
\newcommand{\argmin}{\operatornamewithlimits{argmin}}
\begin{document}
\newcommand{\fwidthtwo}{0.3}
\newcommand{\fwidththree}{0.2}
\newcommand{\fwidthgraph}{0.4}

\makeatletter
\newcommand\figcaption{\def\@captype{figure}\caption}
\newcommand\tabcaption{\def\@captype{table}\caption}
\makeatother

\title{Coupled Learning for Facial Deblur}
\author{Dayong Tian and Dacheng Tao \textit{Fellow, IEEE}\thanks{\copyright 2016 IEEE. Personal use of this material is permitted. Permission from IEEE must be obtained for all other uses, in any current or future media, including reprinting/republishing this material for advertising or promotional purposes, creating new collective works, for resale or redistribution to servers or lists, or reuse of any copyrighted component of this work in other works.}}
\IEEEtitleabstractindextext{
\begin{abstract}
Blur in facial images significantly impedes the efficiency of recognition approaches. However, most existing blind deconvolution methods cannot generate satisfactory results, due to their dependence on strong edges which are sufficient in natural images but not in facial images.  In this paper, we represent a point spread functions (PSF) by the linear combination of a set of pre-defined orthogonal PSFs and similarly, an estimated intrinsic sharp face image (EI) is represented by the linear combination of a set of pre-defined orthogonal face images. In doing so, PSF and EI estimation is simplified to discovering two sets of linear combination coefficients which are simultaneously found by our proposed coupled learning algorithm. To make our method robust to different kinds of blurry face images, we generate several candidate PSFs and EIs for a test image, and then a non-blind deconvolution method is adopted to generate more EIs by those candidate PSFs. Finally, we deploy a blind image quality assessment metric to automatically select the optimal EI. Thorough experiments on the The Facial Recognition Technology (FERET) Database\footnote{\url{http://www.itl.nist.gov/iad/humanid/feret/feret_master.html}}, extended Yale Face Database B\footnote{\url{http://vision.ucsd.edu/~leekc/ExtYaleDatabase/ExtYaleB.html}}, CMU Pose, Illumination, and Expression (PIE) database\footnote{\url{https://www.ri.cmu.edu/research_project_detail.html?project_id=418&menu_id=261}} and Face Recognition Grand Challenge (FRGC) Database version 2.0\footnote{\url{http://www.nist.gov/itl/iad/ig/frgc.cfm}} demonstrate that the proposed approach effectively restores intrinsic sharp face images and consequently improves the performance of face recognition.
\end{abstract}
}
\maketitle
\IEEEdisplaynontitleabstractindextext
\IEEEpeerreviewmaketitle
\section{Introduction}
\label{sec:intro}
Facial blur is common in recorded face images. Examples include motion blur caused by the relative movement between the target and the camera, and out-of-focus blur caused by misalignment between the target and the camera focus. It remains challenging to improve the quality of an observed blurred face image (OB) for subsequent use in various applications, including face recognition and editing. A straightforward method of overcoming OB is to use blind deconvolution methods~\cite{Shan_SIGGRAPH2008}\cite{Fergus_SIGGRAPH2006}\cite{gs1}\cite{gs2}\cite{gs3}\cite{gs4} to obtain an estimated intrinsic face image (EI), and then to exploit the EI for subsequent recognition and analysis. The success of blind deconvolution methods designed for natural images relies on strong edges~\cite{Levin_PAMI2011}, which are relatively rare in most OBs. Therefore, this approach tends to perform poorly~\cite{JRR} and the obtained EIs do not significantly improve subsequent recognition.\\
\indent Recently, machine learning has been exploited to deconvolute OBs. Liao et~al.~\cite{Liao_TIP2005} decomposed an intrinsic sharp face image into the eigen-face subspace and adopted a Gaussian prior to regularizing the EI. However, this approach assumed that the point spread function (PSF) has only one varying parameter, such as a Gaussian kernel with variable variance or a horizontal linear motion function of varying length, and fails to restore images blurred by sophisticated PSFs, e.g., a linear motion function with two varying parameters (direction and length).\\
\indent Nishiyama et~al.~\cite{FADEIN} proposed the FAcial DEblur INference (FADEIN) scheme, which models the PSF estimation procedure as a classification-like problem. By calculating the correlation matrix $\mathbf{A}_i$ that encodes the 2D Fourier transform features of all training images blurred by the $i$-th pre-defined PSF, the subspace $\phi_i$, corresponding to the leading eigenvectors of $\mathbf{A}_i$, is used to model the $i$-th PSF. The PSF corresponding to the subspace closest to the feature of the OB is then exploited to deconvolute the OB. In this way, FADEIN can only model a finite (and, in reality, small) discrete set of PSFs and fails to model PSFs that are not defined in the training stage.\\
\begin{figure}[t]
\centering
\includegraphics[width=0.6\linewidth]{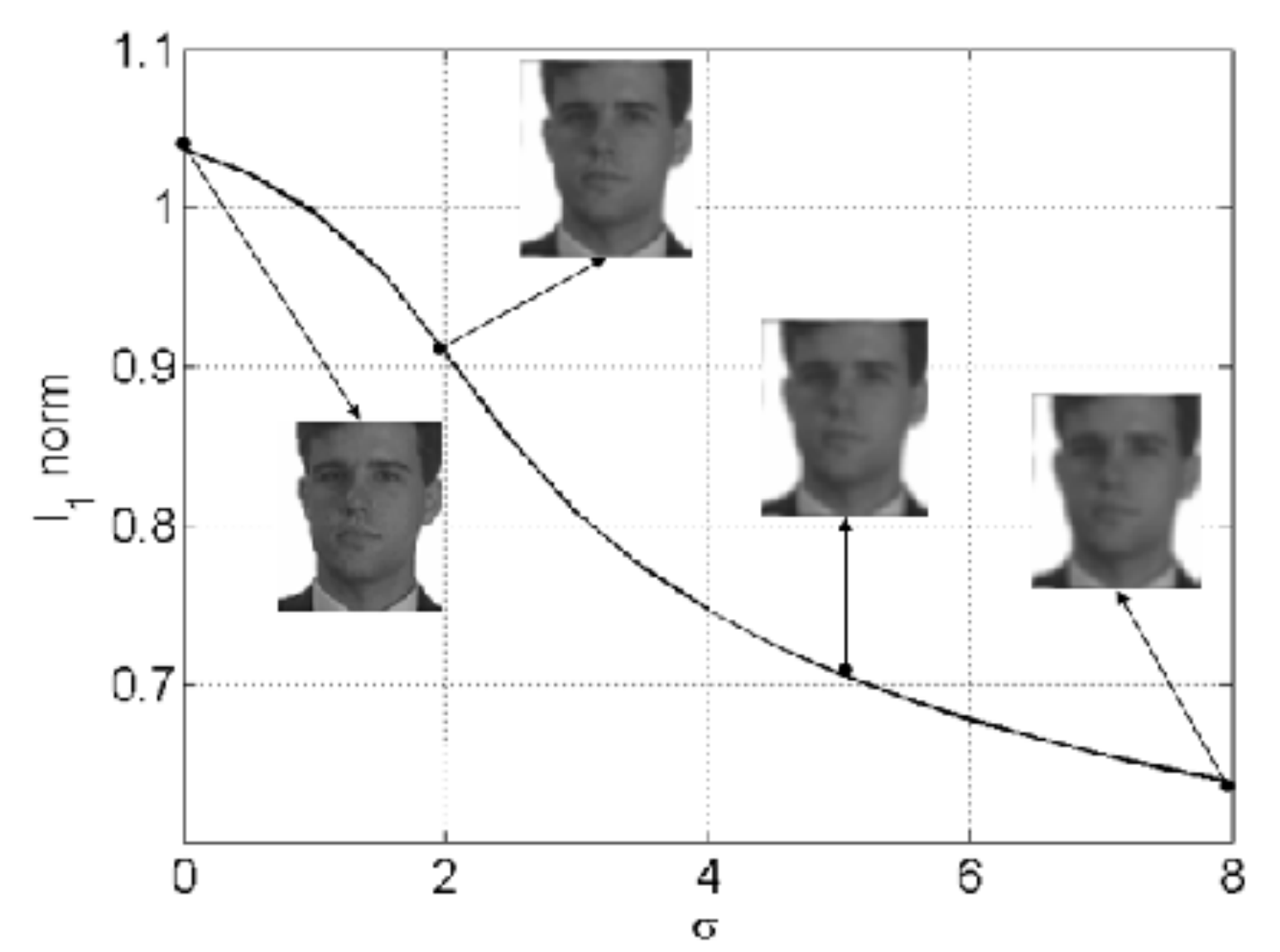}
\caption{Illustration of the drawbacks of the sparse prior. The image is blurred by a Gaussian kernel of standard deviation 0-8. The $l_1$-norm of sparse coefficients monotonously decreases as the standard deviation increases. Hence, the sparse prior may lead to a blurry result.}
\label{fig:drawbackofJRR}
\end{figure}
\indent Zhang et~al.~\cite{JRR} proposed the Joint Restoration and Recognition (JRR) scheme, which combines restoration and recognition within the framework of sparse learning. JRR assumes that the intrinsic sharp face image of an OB can be represented as a linear combination of face images in the training set, and the coefficients of the linear combination are assumed to be sparse. Although sparse prior has been proven to be effective in a wide range of applications including face recognition\cite{jiangyou1}\cite{SparseRecognition1}\cite{SparseRecognition2}\cite{SparseRecognition3} and image restoration\cite{SparseDeblur2}\cite{SparseDeblur1}\cite{SparseDeblur3}\cite{SparseDeblur4}, as shown in Figure~\ref{fig:drawbackofJRR}, it is inappropriate for some deconvolution tasks, since the sparse prior may predispose to blurry images. In this situation, JRR may fail to deconvolute the OB. Furthermore, six parameters need to be empirically tuned, making JRR difficult to use in practice.\\
\indent Recently, J. Pan et al.~\cite{eccvguanshui2014} proposed a face image deblurring method based on the contour of faces.  The useless edges of a face, such as those around eyes and eyebrows, are removed firstly, because these edges have negative effects on the PSF estimation. Then, the PSF is estimated by finding a template for an OB from the training gallery. Rather than utilizing the information on edges intrinsically, like those unsupervised methods, J. Pan et al. try to utilize such information directly and smartly by filtering the edges. However, this work still depends on the edges.\\
\indent To avoid the aforementioned problems and to conduct high performance face restoration, we cast the facial deblur procedure as a regression-like problem based on two mild assumptions: (1) any PSF can be represented by a linear combination of a set of orthogonal PSFs; and (2) the intrinsic sharp face image of an OB can be represented linearly by a set of orthogonalized sharp face images.\\
\begin{figure*}[t]
\centering
\includegraphics[width=0.8\linewidth]{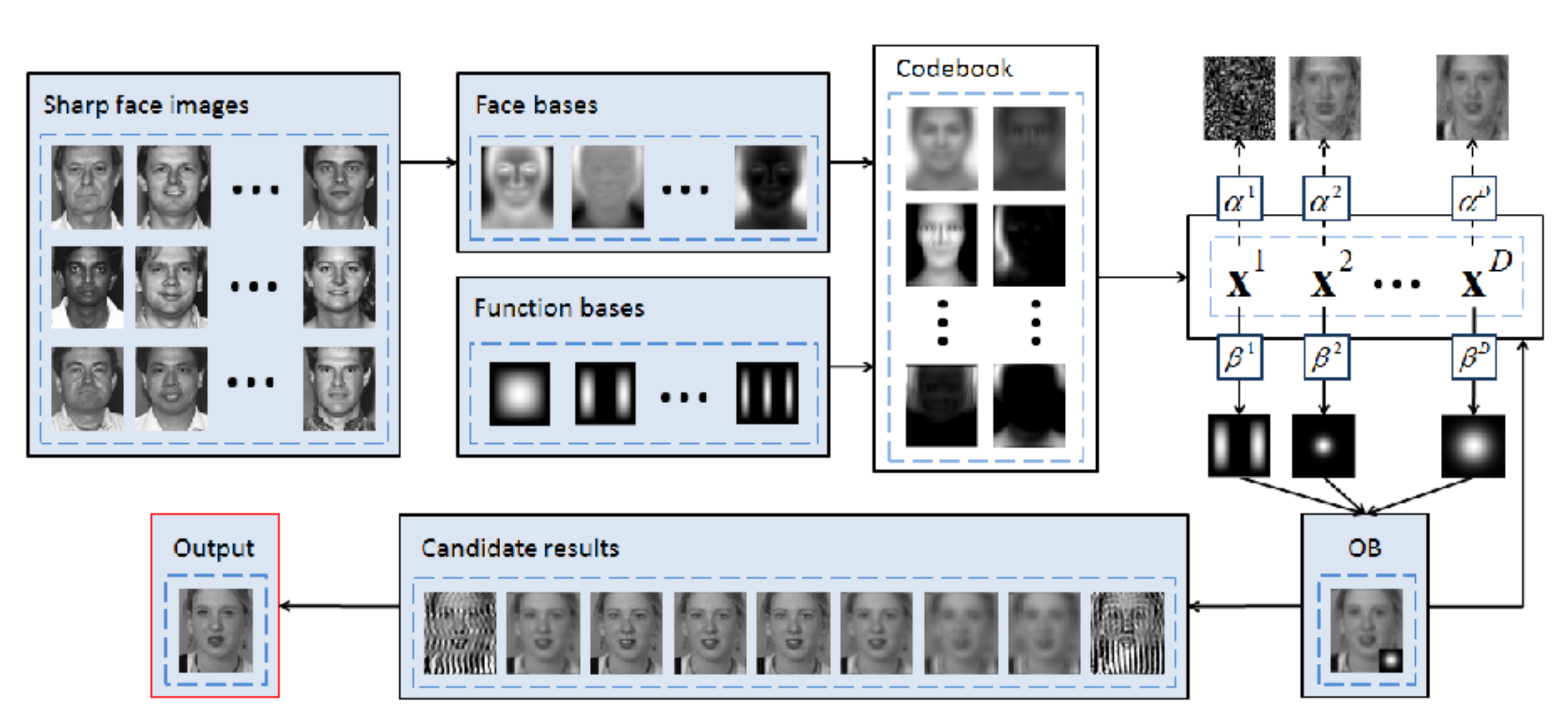}
\caption{The framework of our method.}
\label{fig:framework}
\end{figure*}
\indent Under the above two assumptions, we develop a coupled learning algorithm (shown in Figure~\ref{fig:framework}) to simultaneously calculate all possible PSFs and EIs by discovering the coefficients of two associated linear combinations, in which each PSF corresponds to a particular EI. Empirically, all the EIs show far from satisfactory results, for the following two reasons. First, the dissimilarities between the training sharp face images and the intrinsic sharp face image of the OB can result in reconstruction errors. Second, the parameter space of EI is of tens of thousands of dimensions and thus to obtain a high quality EI requires a large number of training sharp images. By contrast, the parameter space of PSF is much smaller and can be estimated precisely given a small size training set. We therefore generate a sequence of PSFs and use a classical non-blind deconvolution method~\cite{Chan_TIP2011} to deblur the OB and generate candidate results. Lastly, a blind image quality assessment (BIQA) method~\cite{BIQA} is adopted to automatically select the best EI which corresponds to a particular PSF.\\
\indent In contrast to conventional face recognition scheme which consists of a face representation stage and a face matching stage~\cite{ding2014multi}, we propose a new recognition method based on our deblurring procedure. Intuitively, when the sharp face images have the same identity as the OB, the resulting EI is of high quality because these sharp face images are more similar to the intrinsic sharp face image of the OB. We therefore only need to deblur an OB using all the sets of sharp face images, where each set only contains sharp face images of one identity. The identity of the set that produces the best deblurring result is assigned to the OB. In this way, the proposed deblurring method simultaneously deblurs and recognizes the OB.\\
\indent This paper is organized as follows. The proposed deconvolution method is described in Section~\ref{sec:model}. In Section~\ref{sec:fast}, we show how to reduce the computational costs for symmetric PSFs. In Section~\ref{sec:exp}, we demonstrate the effectiveness of the proposed deconvolution and recognition methods. We conclude the paper in Section~\ref{sec:con}.
\section{On Combining Coupled Learning and BIQA for Facial Deblur}
\label{sec:model}
The proposed scheme for facial deblur, shown in Figure~\ref{fig:framework}, is comprised of four major steps: (1) codebook construction; (2) coefficient $\mathbf{x}$ computation; (3) candidate PSF construction and candidate results generation; and (4) the BIQA-based best candidate result selection. This section shows the motivations and details of each step.\\
\indent In general, the relationship between an OB $I$, its intrinsic sharp face image $I^o$, and the corresponding PSF $\phi$ is modeled as
\begin{equation}\label{eq:blurmodel}
I=I^o*\phi+\eta,
\end{equation}
where $\eta $ is the additive noise and $*$ is the convolution operation. We assume $I^o$ can be represented by a linear combination of a set of bases $\{v_i\},i=1,\ldots,M$, and $\phi$ can be represented by a linear combination of a set of functions $\{\phi_j\},j=1,\ldots,N$. Hence, we have
\begin{equation}
I^o*\phi=\left(\sum_{i=1}^M{\alpha_i v_i}\right)*\left(\sum_{j=1}^N{\beta_j\phi_j}\right),
\end{equation}
where $\alpha_i$s and $\beta_j$s are coefficients of the linear combinations. Let $vec(\cdot)$ be the vectorization operation. Let the $(j+(i-1)N)$th column of matrix $\mathbf{A}$ be $vec(v_i*\phi_j)$ and the $(j+(i-1)N)$th element of vector $\mathbf{x}$ be $\alpha_i\beta_j$. Equation~\eqref{eq:blurmodel} can be rewritten as
\begin{equation} \label{eq:ourmodel}
\mathbf{I}=\mathbf{A}\mathbf{x}+vec(\eta).
\end{equation}
where $\mathbf{I}=vec(I)$. In our approach, $\{v_i\}$ and $\{\phi_j\}$ are predefined. Hence, the remaining problems are calculating $\mathbf{x}$ in Equation~\eqref{eq:ourmodel} and calculating $\alpha_i$ and $\beta_j$ from $\mathbf{x}$.
\subsection{Construct $\mathbf{A}$}
Given a set of sharp face images $\{I_p\},p=1,\ldots,P$, we use the first $M$ left singular vectors (i.e., those corresponding to the largest $M$ singular values) of matrix $\mathbf{D}$ as $\mathbf{v}_i$ , where $\mathbf{D}$ is defined as:
\begin{equation}
\mathbf{D} = [\mathbf{I}_1,\mathbf{I}_2,\ldots,\mathbf{I}_p]
\end{equation}
To represent the PSFs, we use a set of orthogonal functions, called function bases
\begin{equation}
\begin{gathered}
f(m,n,x,y)=sin\left(\frac{m\pi x}{a}\right) \sin\left(\frac{n\pi y}{b}\right)\\
m=1,2,\ldots,L,\quad n=1,2,\ldots,L\\
0\leq x\leq a,0\leq y \leq b\\
\end{gathered}
\end{equation}
If the first $K^2(m=n=1,\ldots,K)$ function bases are used, then we let $\phi_{m+(n-1)K}=f(m,n,x,y)$. Hence, we have
\begin{equation}
\begin{split}
{\bf{A}} = [ &vec\left( {{I_1}*{\phi _1}} \right),...,vec\left( {{I_1}*{\phi _{{K^2}}}} \right),...,\\
& vec\left( {{I_P}*{\phi _1}} \right),...,vec\left( {{I_P}*{\phi _{{K^2}}}} \right) ]
\end{split}
\end{equation}
\subsection{Calculate $\mathbf{x}$}
An intuitive way to calculate $\mathbf{x}$ from Equation~\eqref{eq:ourmodel} is by minimizing
\begin{equation}\label{eq:computexintuitive}
\argmin_{\mathbf{x}}\parallel \mathbf{I}-\mathbf{A}\mathbf{x} \parallel^2+R(\mathbf{x}),
\end{equation}
where $R(\mathbf{x})$ is a regularization on $\mathbf{x}$. Here, we do not impose any regularization on $\mathbf{x}$, i.e., $R(\mathbf{x})=0$. Hence, minimizing Equation~\eqref{eq:computexintuitive} is equivalent to solving
\begin{equation}\label{eq:computex}
\mathbf{A}\mathbf{x}=\mathbf{I}.
\end{equation}
Equation~\eqref{eq:computex} has the closed form solution
\begin{equation}
\mathbf{x}=(\mathbf{A}^T\mathbf{A})^{-1}(\mathbf{A}^T \mathbf{I}),
\end{equation}
where $\mathbf{A}^T$ is the transpose of $\mathbf{A}$. Alternatively, the conjugate gradient descent method~\cite{conjgrad} can be used to solve
\begin{equation}\label{eq:cdm}
\mathbf{x^T}(\mathbf{A}^T\mathbf{A})\mathbf{x}-\mathbf{x}^T\mathbf{A}^T \mathbf{I}=0.
\end{equation}
Solving Equation~\eqref{eq:cdm} using the conjugate gradient descent method is much quicker than directly computing the inverse matrix of $\mathbf{A}^T\mathbf{A}$.
\subsection{Calculate $\alpha$ and $\beta$}
Once $\mathbf{x}$ is found, $\alpha$ and $\beta$ can be calculated by solving the following nonlinear equation system:
\begin{equation}\label{eq:computeab}
\begin{gathered}
x_{j+(i-1)N}=\alpha_i\beta_j\\
i=1,\ldots,M\\
j=1,\ldots,N\\
\end{gathered}
\end{equation}
Equation~\eqref{eq:computeab} has $M+N$ unknown variables and $MN$ equations, and is usually an over-determined system (i.e., $MN>M+N$). Hence, it can be solved by minimizing
\begin{equation}\label{eq:computeabintuitive}
\underset{\beta_j,1\leq j\leq N}{\underset{\alpha_i,1\leq i\leq M}{\argmin{}}}\sum_{i=1}^M{\sum_{j=1}^N{(x_{j+(i-1)N}-\alpha_i\beta_j)^2}}.
\end{equation}
Unlike linear over-determined equation systems, where the solution is unique when the rank of the coefficient matrix equals the number of columns, Equation~\eqref{eq:computeabintuitive} may have multiple solutions. For example, if $\{\alpha_i^*\}$ and $\{\beta_j^*\}$ are solutions of Equation~\eqref{eq:computeabintuitive}, $\{\alpha_i^*/c\}$ and $\{c\beta_j^*\}$ will also be solutions, where $c$ is a non-zero constant. Therefore, prior knowledge, regularizations, or constraints are required.\\
\begin{figure}[t]
\centering
\includegraphics[width=0.6\linewidth]{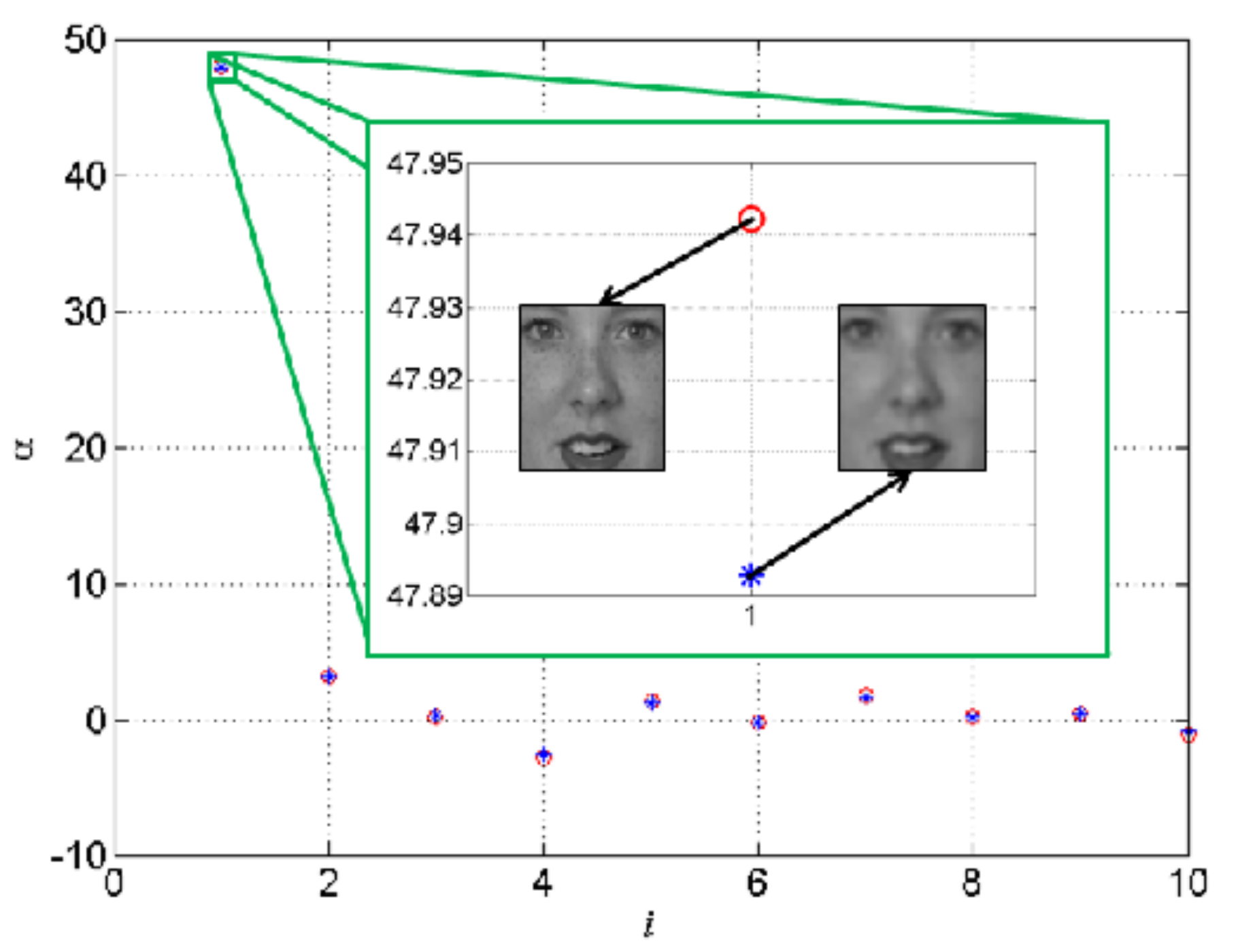}
\caption{The difference between projections of an image and its blurred counterpart. 90\% images in FERET dataset are used to construct ${\bf D}$ and the first 10 projections of a test image in the remaining 10\% images and its blurred counterpart which is blurred by a Gaussian PSF with standard deviation 2 are shown here. It can be found that projections on the first left singular vector of these two images are very close to each other. The difference is approximately 0.1\%.}
\label{fig:alphaprior}
\end{figure}
\indent We observe that the projections of $I^o$ on $\{v_i^*\}$ (i.e., $\alpha_i$s) are similar to those of $I$ (Figure~\ref{fig:alphaprior}). Hence, the following item can be added into Equation~\eqref{eq:computeabintuitive}:
\begin{equation}
(\langle I,v_1 \rangle-\alpha_1)^2,
\end{equation}
where $\langle I,v_1 \rangle$ is the inner product, i.e., the projection of $I$ on $v_1$.\\
It is desirable that the averages of the local windows of an OB should not be altered too much by an estimated PSF. Therefore, its integration should be equal to 1, that is,
\begin{equation}
\sum_{j=1}^N{\beta_j\int_\Omega{\phi_jd\Omega}}=1,
\end{equation}
where $\Omega$ is the domain in which the PSF is defined. The PSF should also be positive, that is,
\begin{equation}
[vec(\phi_1),vec(\phi_2),\ldots,vec(\phi_N)][\beta_1,\beta_2,\ldots,\beta_N]^T\geq [0,0,\ldots,0]^T.
\end{equation}
In conclusion, the proposed minimization problem is
\begin{equation}\label{eq:abfinal}
\begin{gathered}
\underset{\beta_j,1\leq j\leq N}{\underset{\alpha_i,1\leq i\leq M}{\argmin{}}}{\sum_{i=1}^M{\sum_{j=1}^N{(x_{j+(i-1)N}-\alpha_i\beta_j)^2}}+(\langle I,v_1 \rangle-\alpha_1)^2}\\
s.t. \sum_{j=1}^N{\beta_j\int_\Omega{\phi_jd\Omega}}=1\\
[vec(\phi_1),\ldots,vec(\phi_N)][\beta_1,\ldots,\beta_N]^T\geq [0,\ldots,0]^T\\
\end{gathered}
\end{equation}
As SVD is adopted to generate face bases, we only use the projection on the first left singular vector which corresponds to the largest singular value as a regularization in the object function of our optimization problem. The problem~\eqref{eq:abfinal} can be solved using the augmented Lagrange multiplier method~\cite{ALag}.\\
\subsection{Generate candidate results}
Since reconstruction errors exist, the $\alpha_i$s calculated by the procedure above cannot usually reconstruct satisfactory results, especially when the identity of the OB is not included in the training set. However, as stated in the introduction, a PSF is much easier to estimate than its corresponding EI. Setting $M$ as a certain value, one can get a PSF correspondingly. We therefore generate a sequence of PSFs by setting different $M$'s and use a classical non-blind deconvolution method~\cite{Chan_TIP2011} to deblur the OB to generate $M$ candidate results.\\
\begin{figure}[t]
\centering
\includegraphics[width=0.8\linewidth]{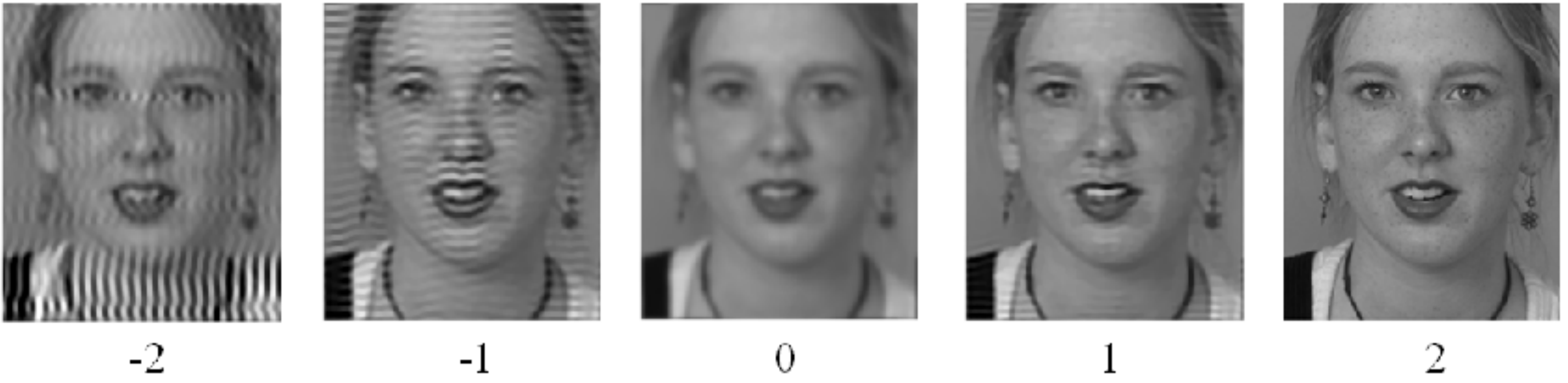}
\caption{Typical results and their corresponding scores. -2 denotes a failure, 0 is a blurry result, and 2 is a sharp image.}
\label{fig:score}
\end{figure}
\subsection{Assessing candidate results}
BIQA across different images is a challenging task. However, blindly assessing the qualities of images distorted by different ways from an image is much easier. Among various BIQA methods, BRISQUE-L~\cite{BIQA} feature is used here to select the best candidate PSF, due to its robustness and computational efficiency.\\
\indent BRISQUE pre-processes an image by local mean removal to generate mean subtracted contrast normalized (MSCN) coefficients which are fitted by Asymmetric Generalized Gaussian Model (AGGD) in~\cite{BRISQUE}. The parameters of AGGD are estimated to form up an 18-dimensional feature. This procedure is done on two scales and results in a 36-dimensional feature. To robustify BRSIQUE, BRISQUE-L introduced L-moments which are closely related to L-estimators and extensively used in robust image filtering theory~\cite{ImageFiltering}. BRISQUE-L is also extracted on two different scales and totally 36-dimensional.
\subsection{Simultaneous restoration and recognition}
\indent A subset of candidate results is manually evaluated at five levels, as shown in Figure~\ref{fig:score}. Finally, support vector regression (SVR) is trained to automatically evaluate the quality of the remaining candidate results.\\
\indent Since a BRISQUE-L feature contains two parts (18 elements in each part), a multiple kernel learning (MKL) method~\cite{MKL} needs to be adopted. Of the various MKL algorithms, the SMO-MKL approach~\cite{SMO-MKL} has been shown to be efficient and effective across a wide range of applications, and its source code is available online\footnote{\url{http://research.microsoft.com/en-us/um/people/manik/code/SMO-MKL/download.html}}. We therefore use it here with five Gaussian kernels for each part and five Gaussian kernels for the whole feature, that is, 15 kernels in total.\\
\indent Let a BRISQUE-L feature be $\mathbf{w}=(w_1,\ldots,w_{36})^T$ and $\mathbf{w}(i:j)$ is a vector comprised of the $i$th to $j$th elements in $\mathbf{w}$. The standard deviation $\sigma$ of Gaussian kernel is chosen from set $\{2^{-2},2^{-1},2^0,2^1,2^2\}$. Hence, the fifteen Gaussian kernels are:
\begin{equation}
\begin{gathered}
G_1=\exp\left(-\frac{\mathbf{w}^T(1:18)\mathbf{w}(1:18)}{(2^{-2})^2}\right),\ldots,\\
G_5=\exp\left(-\frac{\mathbf{w}^T(1:18)\mathbf{w}(1:18)}{(2^{2})^2}\right),\\
G_6=\exp\left(-\frac{\mathbf{w}^T(19:36)\mathbf{w}(19:36)}{(2^{-2})^2}\right),\ldots,\\
G_{10}=\exp\left(-\frac{\mathbf{w}^T(19:36)\mathbf{w}(19:36)}{(2^{2})^2}\right),\\
G_{11}=\exp\left(-\frac{\mathbf{w}^T\mathbf{w}}{(2^{-2})^2}\right),\ldots,\\
G_{15}=\exp\left(-\frac{\mathbf{w}^T\mathbf{w}}{(2^{2})^2}\right)
\end{gathered}
\end{equation}
\indent The primal problem of multiple kernel learning for SVR is
\begin{equation}\label{eq:svrprimal}
\begin{gathered}
\min_{\mathbf{z},b,\mathbf{\xi}^\pm\geq \mathbf{0},\mathbf{d}\geq \mathbf{0}}
{\frac{1}{2}\sum_k{\mathbf{z}_k^T\mathbf{z}_k/d_k}+C\sum_h{(\xi_h^++\xi_h^-)}+\frac{\lambda}{2}
\left(\sum_k{d_k^p}\right)^p}\\
s.t. \pm\left(\sum_k{\mathbf{z}_k^TG_k(\mathbf{w}_h)}+b-s_h\right)\leq \epsilon+\xi_h^\pm
\end{gathered}
\end{equation}
where $\mathbf{z_k}$ is the support vector and $d_k$ is the kernel weights of the linear combination of base kernels $\{G_k\}$. $\xi_h^+$ and $\xi_h^-$ are slack variables allowing for errors around the regression function. $C$, $\lambda$, $p$ and $\epsilon$ are positive constants set empirically. Introducing Lagrange Multipliers $a_h^+$ on constraints corresponding to $\xi_h^+$ and $a_h^-$ on constraints corresponding to $\xi_h^-$, the dual problem of Eq.~\eqref{eq:svrprimal} is
\begin{equation} \label{eq:svrdual}
\begin{aligned}
\underset{\mathbf{0}\leq \mathbf{a}^+,\mathbf{a}^-\leq C\mathbf{1}}{\underset{\mathbf{1}^T\mathbf{a}^+=\mathbf{1}^T\mathbf{a}^-}{\max{}}}
\mathbf{1}^T(S(\mathbf{a}^+-\mathbf{a}^-)-\epsilon(\mathbf{a}^++\mathbf{a}^-))-\\
\frac{1}{8\lambda}
\left(\sum_k{\left((\mathbf{a}^+-\mathbf{a}^-)^TG_k(\mathbf{a}^+-\mathbf{a}^-)\right)^q}\right)^\frac{2}{q}
\end{aligned}
\end{equation}
where $S$ is a diagonal matrix whose elements are the scores of training samples. $p$ and $q$ satisfy
\begin{equation}
\frac{1}{p}+\frac{1}{q}=1
\end{equation}
Sequential Minimal Optimization (SMO) \cite{smo1}\cite{smo2}\cite{smo3} is used to solve Eq.~\eqref{eq:svrdual}.
\indent The proposed recognition method is based on the SVR outputs. We construct $\mathbf{A}$ for each identity, which is then used to deblur each OB. Lastly, the identity of $\mathbf{A}$ that produces the best deblurring result is assigned to the OB. This simple manipulation results in the simultaneous production of deblurring and recognition results.
\section{Efficient Implementation for Symmetric PSF}
\label{sec:fast}
In the theoretical point view, our algorithm can inherently handle any kinds of PSF. It is based on the mathematical theory - any bounded function can be represented by a linear combination of a complete set of orthogonal functions.\\
\indent We mainly focus on three types of blurring: out-of-focus blur (approximated by a Gaussian kernel), linear motion blur, and a combination of the two. Here, we show how to reduce computational costs by considering the symmetry of these three types of blur. The aim is to reduce the number of function bases, i.e., $N$.\\
\begin{figure}[t]
\centering
\includegraphics[width=0.9\linewidth]{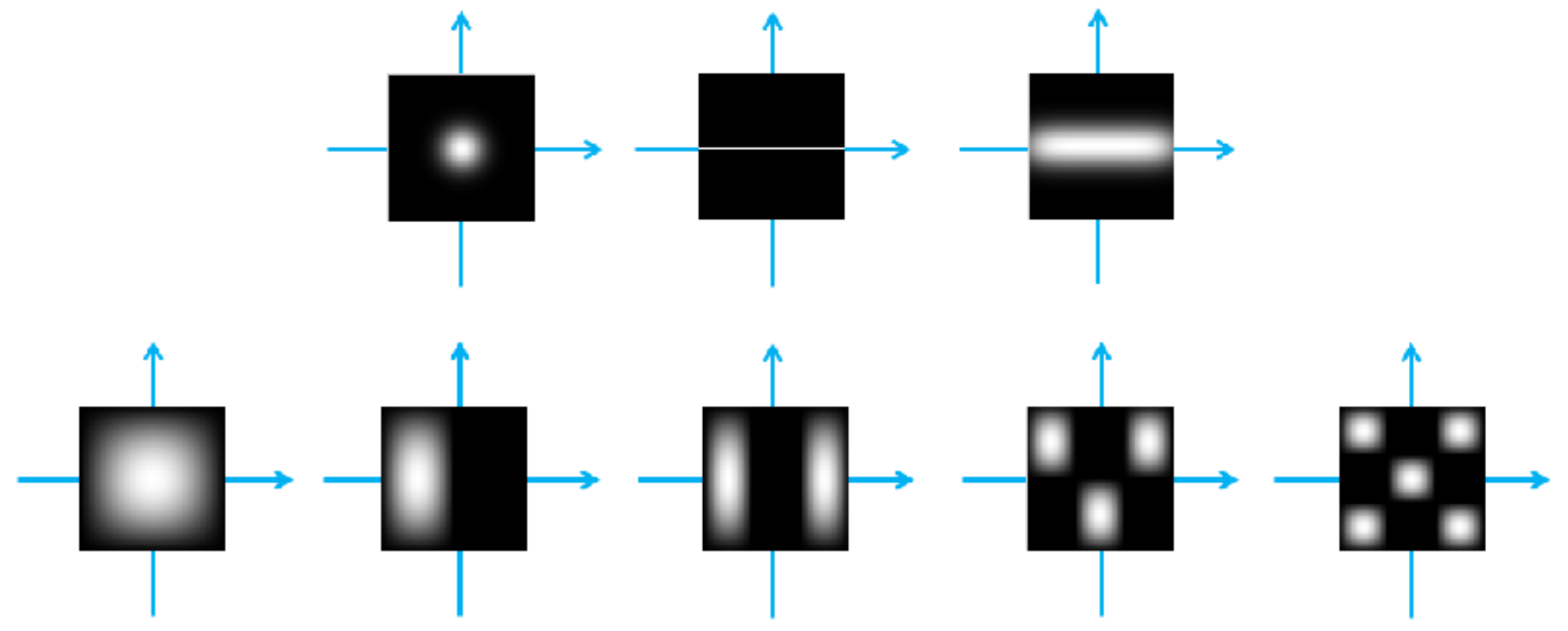}
\caption{The symmetry of PSFs. The top row illustrates three PSFs: a Gaussian kernel, a linear motion kernel, and the combination of both. The bottom row illustrates five function bases, i.e.,$m=n=1$, $m=1\& n=2$, $m=1\& n=3$, $m=2\&n=3$, and $m=n=3$. The function bases of odd $m$ and $n$ are symmetric about the two axes and can be used to represent the three symmetric PSFs.}
\label{fig:symmetry}
\end{figure}
\indent To simplify our analysis, we assume that the PSFs of the linear motion blur and the combined blur only have four directions: $0,\pi/4,\pi/2$ and $3\pi/4$. Considering the symmetry of $0$-PSFs (Figure~\ref{fig:symmetry}), only the $f(m,n,x,y)$s that are symmetric about $x=a/2$ and $y=b/2$ (i.e., $m=1,3,5,\ldots$ and $n=1,3,5,\ldots$), can be used to represent such PSFs. Since both the symmetric axes of $0$-PSFs and $\pi/2$-PSFs are $x=a/2$ and $y=b/2$, they can be represented by a common set of function bases, $\{\varphi_j\}$. By rotating this set of function bases by $\pi/4$ clockwise or counter-clockwise, the function bases set (denoted as $\{\psi_j\}$) can reconstruct $\pi/4$-PSFs and $3\pi/4$-PSFs.\\
\indent Therefore, if the directions of the PSFs of OBs can be estimated, the OBs can be divided into two groups: Group I($0$- \& $\pi/2$-PSFs) and Group II ($\pi/4$ \& $3\pi/4$-PSFs). By constructing two different $\mathbf{A}$s using $\{\varphi_j\}$ and $\{\psi_j\}$, respectively, an OB can be deconvoluted using the method proposed in Section 2. The only remaining problem is how to efficiently estimate the direction of the PSFs.\\
\begin{figure}[t]
\centering
\includegraphics[width = 0.6\linewidth]{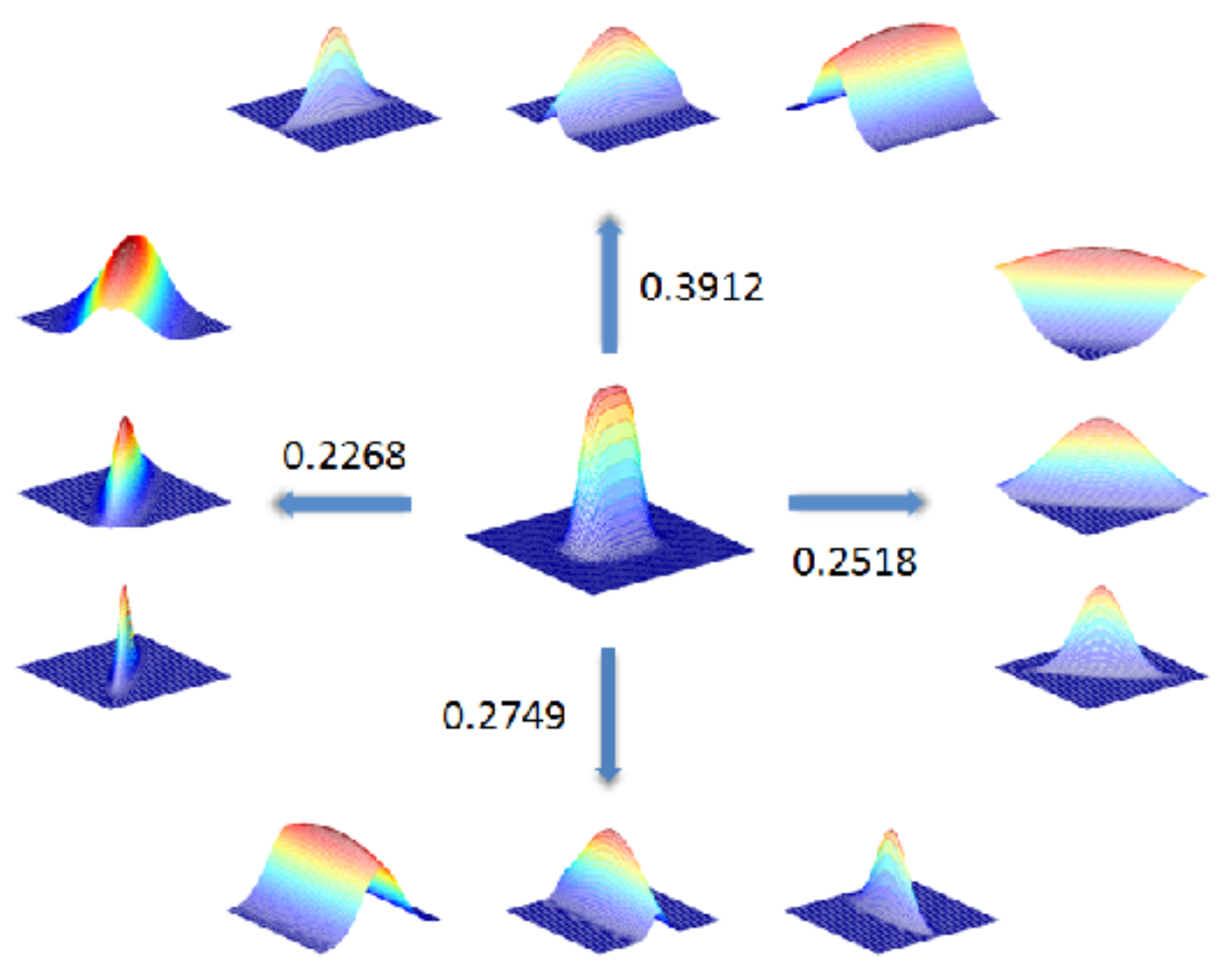}
\caption{Illustration of proposed PSF direction estimation algorithm. A combination of linear motion and Gaussian PSF locates at the center of the graph. $\phi'$s with different $\sigma$ locate around the PSF according to their direction. The numbers are the maximum values of $\int_\Omega{\arrowvert\mathcal{F}(\phi')\arrowvert\arrowvert\mathcal{F}(\phi)\arrowvert}d\Omega$ on each direction.}
\label{fig:direction}
\end{figure}
\subsection{PSF direction estimation}
Ignoring additive noise and taking the Fourier transform on both sides of Equation~\eqref{eq:blurmodel}, we get
\begin{align}
\mathcal{F}(I)=\mathcal{F}(I^o)\mathcal{F}(\phi),
\arrowvert \mathcal{F}(I) \arrowvert=\arrowvert \mathcal{F}(I^o) \arrowvert \arrowvert \mathcal{F}(\phi) \arrowvert,
\end{align}
where $\mathcal{F}(\cdot)$ denotes the Fourier transform. Given another PSF $\phi'$, it can be deduced that
\begin{equation} \label{eq:sb}
\begin{aligned}
2\arrowvert \mathcal{F}(I)\arrowvert \arrowvert \mathcal{F}(\phi')\arrowvert = 2\arrowvert \mathcal{F}(I^o)\arrowvert \arrowvert \mathcal{F}(\phi)\arrowvert \arrowvert \mathcal{F}(\phi')\arrowvert \\
\leq\arrowvert \mathcal{F}(I^o)\arrowvert (\arrowvert \mathcal{F}(\phi)\arrowvert^2+\arrowvert \mathcal{F}(\phi')\arrowvert^2).
\end{aligned}
\end{equation}
\indent In inequality~\eqref{eq:sb}, the equation holds, if and only if $\arrowvert \mathcal{F}(\phi) \arrowvert = \arrowvert \mathcal{F}(\phi')\arrowvert$. Inspired by the correlation-based shape alignment method~\cite{shapealginment}, we try several $\phi'$s that are similar to $\phi$ and find the $\phi'$ whose $\arrowvert \mathcal{F}(\phi')\arrowvert$ maximizes $\int_\Omega{2\arrowvert \mathcal{F}(I^o)\arrowvert \arrowvert \mathcal{F}(\phi)\arrowvert \arrowvert \mathcal{F}(\phi')\arrowvert}$ (Fig.~\ref{fig:direction}). Here, we use following function as $\phi'$:
\begin{equation}
\phi'(x,y,\gamma,\sigma,\theta)=\exp{\left(\frac{x'^2+\gamma^2y'^2}{\sigma^2}\right)},\quad \gamma\geq 1,
\end{equation}
where $x'=x\cos\theta+y\sin\theta$ and $y'=y\cos\theta-x\sin\theta$. \\
\indent The proposed direction estimation method is shown in \textbf{Algorithm 1}. Although we neglect the additive noises to develop our direction estimation algorithm, the algorithm turns out to be efficient when noises exist (Subsection~\ref{subsec:PSFdirect}). A reasonable explanation is that noises are random and directionless, and hence have little effect on direction estimation.
{\renewcommand\baselinestretch{1.5}\selectfont
\begin{algorithm}[htbp]
\caption{~~Direction Estimation Method}
\label{alg:1}
\begin{algorithmic}[1]
\State \textbf{Input:} $I$
\For{each $\theta\in\{0,\pi/4,\pi/2,3\pi/4\}$}
\For{each $\sigma\in\{1,2,\ldots,9\}$}
\If{$V<\arrowvert \mathcal{F}(I) \arrowvert \arrowvert \mathcal{F}(\phi')\arrowvert$}
\State $V=\arrowvert \mathcal{F}(I) \arrowvert \arrowvert \mathcal{F}(\phi')\arrowvert$
\State $L=\theta$
\EndIf
\EndFor
\EndFor
\State \textbf{Output:} $L$
\end{algorithmic}
\end{algorithm}}
\section{Experiments}
\label{sec:exp}
The aim of the following experiments is three-fold: (1) to verify the accuracy of the proposed direction estimation method; (2) to validate the deconvolution method without recognition; and (3) to test the proposed recognition method.\\
\indent We conducted experiments on FERET~\cite{FERET}, CMU-PIE~\cite{CMU-PIE} and extended Yale B~\cite{eYaleB} datasets and compared our method with FADEIN~\cite{FADEIN}, Krishnan~\emph{et al.} method~\cite{Krishnan_CVPR2011}, JRR~\cite{JRR}and Pan~\emph{et~al}~\cite{eccvguanshui2014}.\\
\indent \textbf{Dataset description.} For FERET, we used the gallery (FA set) containing 1,196 images of 1,196 subjects. For CMU-PIE, there are 67 identities and we collected the images named as ``27\_**.jpg'' under the ``illum'' folder of each identity. The file name ``27\_**.jpg'' implies the photo was taken in frontal view. For the extended Yale B, we directly applied the images in ``Cropped Yale'' available at the official website. This subset contains 38 identities and 64 photos taken from the frontal view under different illumination conditions for each identity. For FRGC~2.0, we collect 2000 controlled still images and 500 uncontrolled still images of 200 subjects from both the training and validation partitions. Please note only placid faces were collected and a few uncontrolled still images contain no obvious blurs.\\
\indent \textbf{Evaluation method.} There are three widely used indexes to evaluate image quality: Peak Signal-to-Noise Ratio (PSNR), Universal Image Quality Index (UIQI)~\cite{uiqi_ssim} and Structural Similarity Index (SSIM)~\cite{uiqi_ssim}. PSNR is based on the mean square error between two images, which is easy to calculate but less efficient than indexes based on human visual system~\cite{fourindexes}, such as UIQI and SSIM. Hence, SSIM which is an improvement of UIQI is adopted here to evaluate our deblur method.\\
\indent The SSIM has two inputs, the sharp image and its synthetically blurred counterpart. To calculate SSIM, two means, two standard deviations and one covariance are computed on each $B\times B$ local window of two images as in Eq.~\eqref{eq:SSIM1}.
\begin{equation}\label{eq:SSIM1}
\begin{gathered}
\mu_\zeta=\frac{1}{T}\sum_{q=1}^T{\zeta_q}\qquad \mu_\vartheta=\frac{1}{T}\sum_{q=1}^T{\vartheta_q}\\
\sigma_\zeta^2=\frac{1}{T-1}\sum_{q=1}^T{(\zeta_q-\overset{-}{\zeta})^2}\quad
\sigma_\vartheta=\frac{1}{T-1}\sum_{q=1}^T{(\vartheta_q-\overset{-}{\vartheta})^2}\\
\sigma_{\zeta\vartheta}^2=\frac{1}{T-1}\sum_{q=1}^{T}{(\zeta_q-\overset{-}{\zeta})(\vartheta_q-\overset{-}{\vartheta})}\\
SSIM(\zeta,\vartheta)=\frac{(2\mu_\zeta\mu_\vartheta+c_1)(2\sigma_{\zeta\vartheta}+c_2)}
{(\mu_\zeta^2+\mu_\vartheta^2+c_1)(\sigma_\zeta^2+\sigma_\vartheta^2+c_2)}
\end{gathered}
\end{equation}
where $\zeta$ and $\vartheta$ represent local windows on sharp images and blurred images, respectively. $c_1$ and $c_2$ are two constants. $T=B^2$ is the total number of a local window. The overall SSIM index is the mean of $SSIM(\zeta,\vartheta)$ of all local windows. In our experiments, the mean ($\mu$) and standard deviation ($\sigma$) of SSIM are given on every dataset.\\
\indent \textbf{Global settings.} Throughout our experiments, 30 dB additive white noise is synthesized into test images. We use the default setting of Chan~\emph{et~al.}'s non-blind deconvolution method. Its code is available online\footnote{\url{http://videoprocessing.ucsd.edu/~stanleychan/deconvtv}}.
\begin{figure*}[t]
\centering
\includegraphics[width=0.8\linewidth]{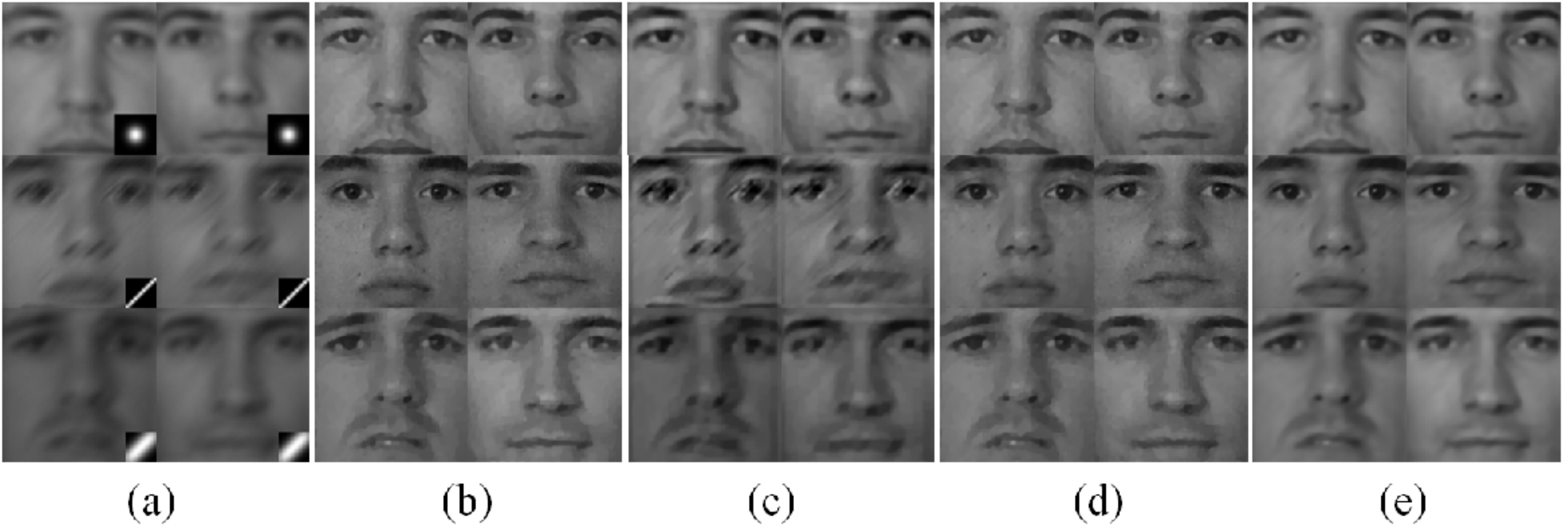}
\caption{Results of experiments on the subset of FERET dataset. (a) Test images and their PSFs; (b) results of the proposed method; (c) results using the method proposed by Krishnan \emph{et~al.}~\cite{Krishnan_CVPR2011}; (d) results of FADEIN~\cite{FADEIN} and (e) results using the method proposed by Pan~\emph{et~al.}~\cite{eccvguanshui2014}}
\label{fig:expFERET}
\end{figure*}
\begin{table*}[b]
\begin{center}
\caption{SSIM Index on FERET}
\label{tb:SSIM_FERET}
\begin{tabular}{|c|c|c|c|c|c|c|}
\hline
&\multicolumn{3}{|c|}{$\mu$}&\multicolumn{3}{|c|}{$\sigma$}\\
\hline
PSF&Combine&Gaussian&Motion&Combine&Gaussian&Motion\\
\hline
FADEIN~\cite{FADEIN}&0.8808&0.9290&0.9053&0.0100&0.0110&0.0123\\
\hline
Krishnan \emph{et~al.}~\cite{Krishnan_CVPR2011}&0.7730&0.8866&0.7177&0.0083&0.0184&0.0324\\
\hline
Pan~\emph{et~al.}~\cite{eccvguanshui2014}&0.8854&0.9211&0.9123&0.0101&0.0103&0.0056\\
\hline
Ours&0.8939&0.9214&0.9158&0.0100&0.0098&0.0043\\
\hline
\end{tabular}
\end{center}
\end{table*}
\subsection{PSF direction estimation}
\label{subsec:PSFdirect}
We randomly selected 1000 images from the three subsets of FERET, CMU-PIE and extended Yale B. The widths of the linear motion kernels were chosen as 3, 5, 7, and 9 and the standard deviations of the Gaussian kernels as 1, 3, 5, 7, and 9. Hence, $4(directions)\times 4(widths) \times 5(standard\quad deviations) \times 1000=80000$ images were generated for testing the proposed direction estimation method. For each combination of these two types of kernels, the rate of correctly estimating the PSF was recorded and shown to be 97\%. Since the direction estimation is imperfect, some OBs will be miscategorized. Hence, OBs in which all candidate results have low scores, say less than 0, should be re-categorized into another group.
\begin{figure*}[t]
\centering
\includegraphics[width=0.8\linewidth]{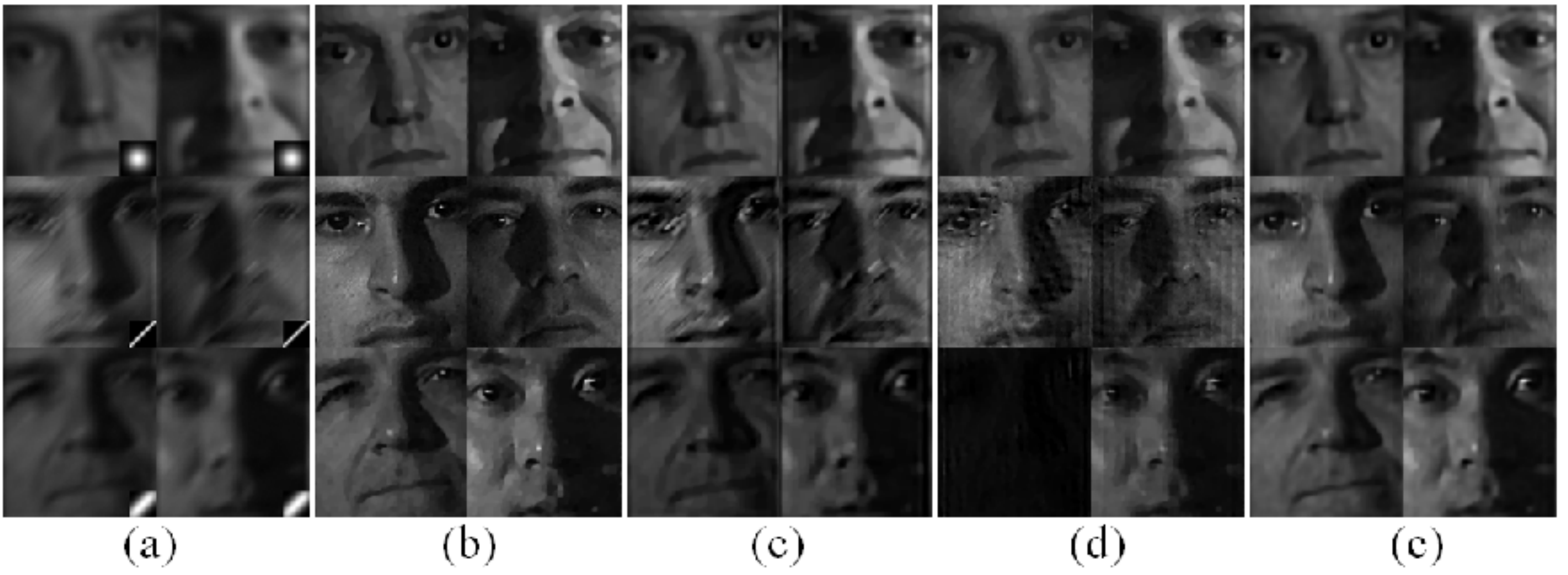}
\caption{Results of experiments on the subset of CMU-PIE dataset. (a) Test images and their PSFs; (b) results of the proposed method; (c) results using the method proposed by Krishnan \emph{et~al.}~\cite{Krishnan_CVPR2011}; (d) results of FADEIN~\cite{FADEIN} and (e) results using the method proposed by Pan~\emph{et~al.}~\cite{eccvguanshui2014}}
\label{fig:expPIE}
\end{figure*}
\begin{figure*}[t]
\centering
\includegraphics[width=0.8\linewidth]{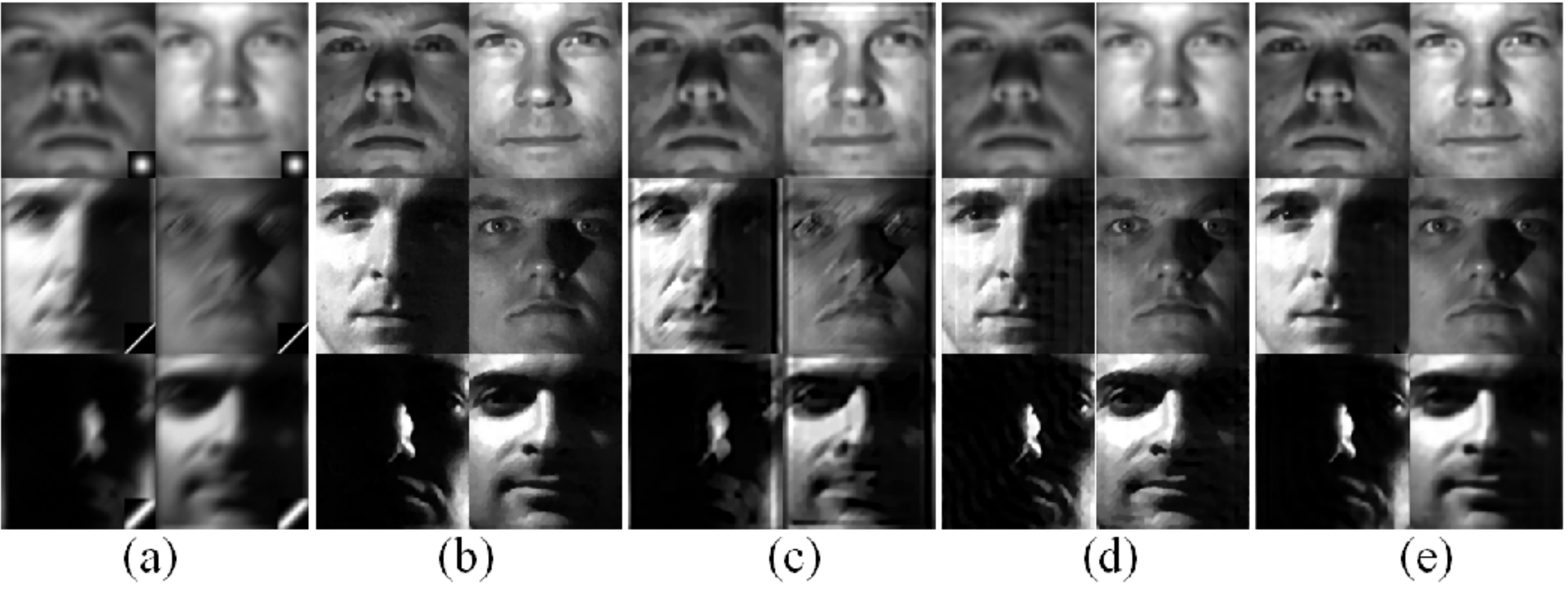}
\caption{Results of experiments on the subset of extended Yale B dataset. (a) Test images and their PSFs; (b) results of the proposed method; (c) results using the method proposed by Krishnan \emph{et~al.}~\cite{Krishnan_CVPR2011}; (d) results of FADEIN~\cite{FADEIN} and (e) results using the method proposed by Pan~\emph{et~al.}~\cite{eccvguanshui2014}}
\label{fig:expyaleb}
\end{figure*}
\begin{table*}[b]
\begin{center}
\caption{SSIM Index on CMU-PIE}
\label{tb:SSIM_PIE}
\begin{tabular}{|c|c|c|c|c|c|c|}
\hline
&\multicolumn{3}{|c|}{$\mu$}&\multicolumn{3}{|c|}{$\sigma$}\\
\hline
PSF&Combine&Gaussian&Motion&Combine&Gaussian&Motion\\
\hline
FADEIN~\cite{FADEIN}	&0.7599    &0.7909    &0.8200			    &0.0171    &0.0174    &0.0224\\
\hline
Krishnan \emph{et~al.}~\cite{Krishnan_CVPR2011}	&0.6398    &0.7705    &0.6579				&0.0212    &0.0171    &0.0463\\	
\hline
Pan~\emph{et~al.}~\cite{eccvguanshui2014}&0.7947&0.8448&0.8033&0.0144&0.0189&0.0151\\
\hline		
Ours	&0.8068    &0.8780    &0.8386				&0.0131    &0.0160    &0.0063\\		
\hline
\end{tabular}
\end{center}
\end{table*}

\begin{table*}[b]
\begin{center}
\caption{SSIM Index on Extended Yale B}
\label{tb:SSIM_yaleb}
\begin{tabular}{|c|c|c|c|c|c|c|}
\hline
&\multicolumn{3}{|c|}{$\mu$}&\multicolumn{3}{|c|}{$\sigma$}\\
\hline
PSF&Combine&Gaussian&Motion&Combine&Gaussian&Motion\\
\hline
FADEIN~\cite{FADEIN}	&0.8458    &0.8537    &0.8537				&0.0349    &0.0547    &0.0547\\	
\hline
Krishnan \emph{et~al.}~\cite{Krishnan_CVPR2011}	&0.6906    &0.8822    &0.6566				&0.0663    &0.0297    &0.0693\\	
\hline
Pan~\emph{et~al.}~\cite{eccvguanshui2014}&0.8521&0.8992&0.8841&0.0334&0.0399&0.0372\\
\hline		
Ours	&0.8998    &0.9167    &0.9020				&0.0384    &0.0516    &0.0221\\				
\hline
\end{tabular}
\end{center}
\end{table*}
\subsection{Facial deblur}
Experiments in this subsection were conducted on FERET. We used 90\% of the images to generate the matrix $\mathbf{A}$. The remaining 10\% of images were treated as OBs and blurred by a Gaussian kernel of $\sigma$ 2, linear motion of length 15 and direction $\pi/4$, or the combination of the two. The first 9 odd order orthogonal functions were used to construct $\mathbf{A}$. We generated nine candidate results for each OB by setting nine $M$s, where the cumulative energy content for the $M$th eigenvector occupied 60\%, 70\%, 80\%, 90\%, 91\%, 92\%, 93\%, 94\% and 95\% of the total energy, respectively. Candidate results of 10 randomly-selected OBs were used to train SVR. FADEIN~\cite{FADEIN} and the Krishnan~\emph{et~al.} method~\cite{Krishnan_CVPR2011} were implemented and examined under the author-suggested settings to guarantee fair comparison. The final results are shown in Figure~\ref{fig:expFERET}. The Krishnan~\emph{et~al.} method did not perform well on this dataset, because of its requirement of strong edges. Since FADEIN is a classification-like scheme, its deconvolution results are similar to ours if the PSFs of the OBs are pre-defined in the training set (the first row). When the PSFs are not pre-defined in the training set (the second and third rows), it tends to give the closest PSFs. Hence, the results are degraded somewhat.
\begin{figure*}[t]
\centering
\includegraphics[width=0.8\linewidth]{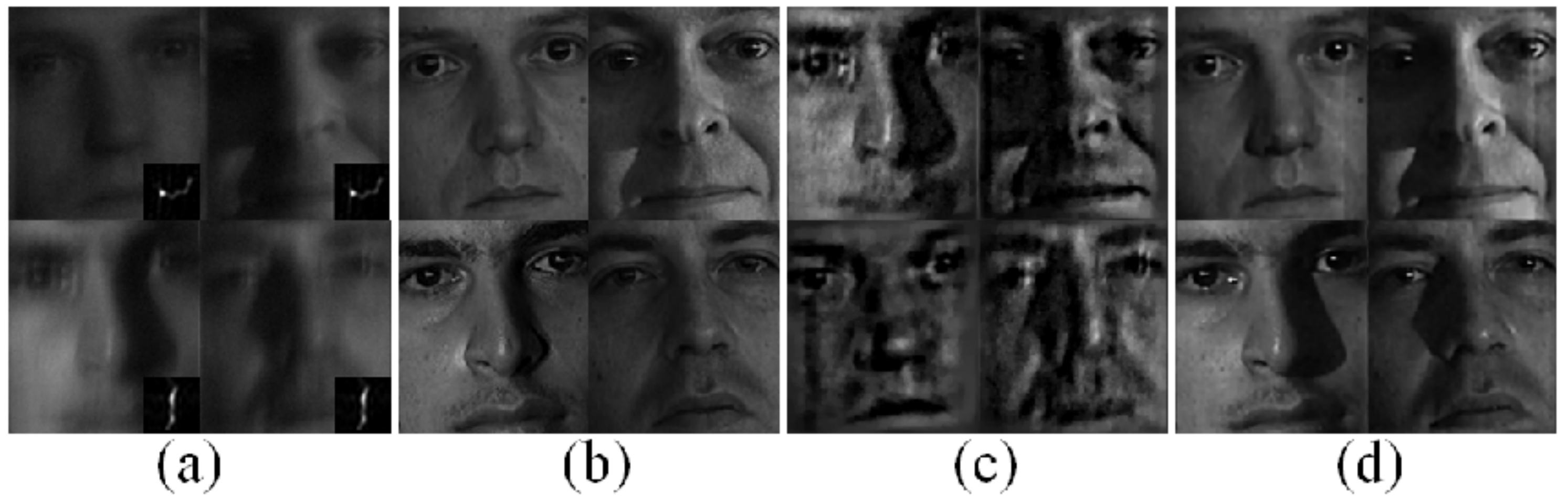}
\caption{Results of experiments on the subset of CMU-PIE dataset. (a) Test images and their PSFs; (b) results of the proposed method; (c) results using the method proposed by Krishnan \emph{et~al.}~\cite{Krishnan_CVPR2011} and (d) results using the method proposed by Pan~\emph{et~al.}~\cite{eccvguanshui2014}}
\label{fig:deblur_camera_pie}
\end{figure*}
\begin{figure*}[t]
\centering
\includegraphics[width=0.8\linewidth]{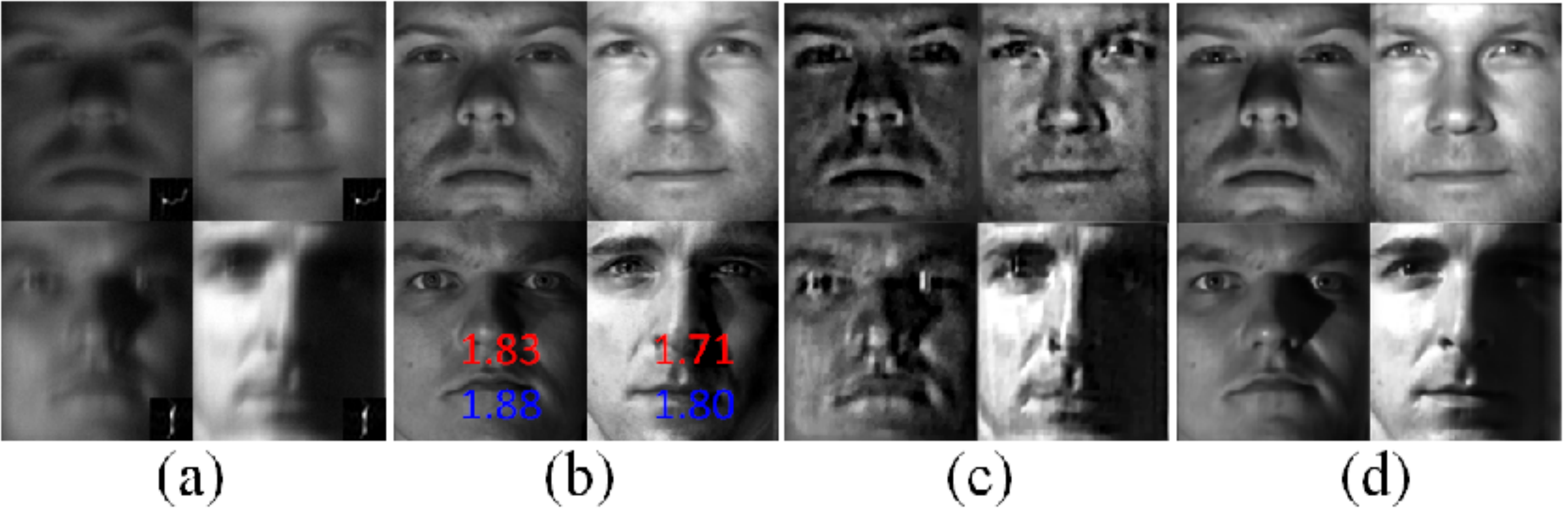}
\caption{Results of experiments on the subset of extended Yale B dataset. (a) Test images and their PSFs; (b) results of the proposed method; (c) results using the method proposed by Krishnan \emph{et~al.}~\cite{Krishnan_CVPR2011} and (d) results using the method proposed by Pan~\emph{et~al.}~\cite{eccvguanshui2014}}
\label{fig:deblur_camera_yale}
\end{figure*}
\subsection{Simultaneous facial deblur and recognition}
\label{subsec:rere}
The proposed method was compared with FADEIN~\cite{FADEIN} and JRR~\cite{JRR} on the subset of CMU-PIE and the subset of the extended Yale B. In contrast to FERET, the face images in these two subsets were taken under different illumination conditions.\\
\indent For experiments on each subset, we adopted the first 50\% of images of each identity in our method for training and the remaining 50\% of images of each identity for testing. All test images were blurred by a Gaussian kernel of $\sigma$ 3, a linear motion of length 15 and direction $\pi/4$ or the combination of both. All these blurred images were collected together as OBs. The first 9 odd order orthogonal functions were used to construct $\mathbf{A}$. The candidate results were generated by setting $M$s in a similar way to those given in Subsection 4.2, but note that $M$ was occasionally zero, because only tens of sharp face images were available to construct matrix $\mathbf{A}$ in each procedure of generating candidate results. We started the procedure with the smallest non-zero .\\
\indent For fair comparison, we used exactly the same setting for our method on each dataset. According to~\cite{FADEIN}, FADEIN adopted the local phase quantization (LPQ)~\cite{LPQ_ICPR} method for recognition. We carefully tuned the parameters of JRR to ensure it reached its best performance.\\
\indent The deconvolution results on CMU-PIE are shown in Figure~\ref{fig:expPIE} and the corresponding recognition rates are listed in Table~\ref{tb:expPIE}. The deconvolution and recognition results on extended Yale B are shown in Fig~\ref{fig:expyaleb} and Table~\ref{tb:expyaleb}, respectively. The method of Krishnan \emph{et~al.}~\cite{Krishnan_CVPR2011} failed to provide satisfactory deconvolution results. Due to the complex illumination conditions, FADEIN~\cite{FADEIN} cannot usually estimate the PSFs precisely and therefore performed poorly on recognition. The deconvolution results of JRR~\cite{JRR} are not shown here, because the authors explained that the deconvolution procedure in JRR was not designed for human visual perception~\cite{JRR}. It is unfair to compare the deconvolution results of JRR with other methods, but note that JRR performed poorly in terms of visual appearance. It can be concluded that the proposed method not only gives satisfactory deconvolution results, but also significantly boosts recognition performance.\\

\indent Since recognition is based on image quality, our proposed method does not require compensation for illumination. However, due to the complex illumination conditions in some images, the proposed method may fail to deconvolute in some cases. Also, since Equation~\eqref{eq:abfinal} has multiple local minima, this can result in erroneous recognition, because the correct $\{\beta_j\}$ can be given even when $\{\alpha_i\}$ is wrong. Even though the probability of this occurring is low, as shown by the high recognition rate, such mistakes are inevitable.\\
\begin{table}[h]
\begin{center}
\caption{Recognition rates on the subset of CMU-PIE dataset.}
\label{tb:expPIE}
\begin{tabular}{|c|c|c|c|}
\hline\
 & Gaussian & Motion & Both\\
\hline
FADEIN+LPQ~\cite{FADEIN}& 85.6 & 84.1 & 82.3\\
\hline
JRR~\cite{JRR} & 92.7 & 92.3 & 91.7\\
\hline
Ours & 95.1 & 95.0 & 95.0\\
\hline
\end{tabular}
\end{center}
\end{table}
\begin{table}[h]
\begin{center}
\caption{Recognition rates on the subset of extended Yale B dataset.}
\label{tb:expyaleb}
\begin{tabular}{|c|c|c|c|}
\hline\
 & Gaussian & Motion & Both\\
\hline
FADEIN+LPQ~\cite{FADEIN} & 77.3 & 75.8 & 70.1\\
\hline
JRR~\cite{JRR} & 88.2 & 86.8 & 86.3\\
\hline
Ours & 93.4 & 92.6 & 91.8\\
\hline
\end{tabular}
\end{center}
\end{table}

\begin{table*}[b]
\begin{center}
\caption{SSIM Index on CMU-PIE dataset}
\begin{tabular}{|c|c|c|c|c|}
\hline
& \multicolumn{2}{|c|}{$\mu$} & \multicolumn{2}{|c|}{$\sigma$} \\
\hline
PSF & TYPE I & TYPE II & TYPE I & TYPE II\\
\hline
Krishnan \emph{et~al.}~\cite{Krishnan_CVPR2011} & 0.4973 & 0.5982 & 0.0321 & 0.0792\\
\hline
Pan~\emph{et~al.}~\cite{eccvguanshui2014}&0.8149&0.7922&0.0930&0.0958\\
\hline
Ours & 0.8566  & 0.8315 & 0.0749 & 0.1834 \\
\hline
\end{tabular}
\end{center}
\end{table*}
\begin{table*}[b]
\begin{center}
\caption{SSIM Index on Extended Yale B Dataset}
\begin{tabular}{|c|c|c|c|c|}
\hline
& \multicolumn{2}{|c|}{$\mu$} & \multicolumn{2}{|c|}{$\sigma$} \\
\hline
PSF & TYPE I & TYPE II & TYPE I & TYPE II\\
\hline
Krishnan \emph{et~al.}~\cite{Krishnan_CVPR2011} & 0.5379 & 0.6428 & 0.0712 & 0.0924\\
\hline
Pan~\emph{et~al.}~\cite{eccvguanshui2014}&0.8514&0.8112&0.0897&0.0926\\
\hline
Ours & 0.8756  & 0.8531 & 0.1273 & 0.1857 \\
\hline
\end{tabular}
\end{center}
\end{table*}
\subsection{Experiments on camera-shaking blur}
\label{subsec:csb}
In this experiment, the restoration results of our method are compared with Krishnan \emph{et~al.} and its recognition results are compared with JRR on CMU-PIE and extended Yale B datasets. FADEIN is not available for this kind of blur, i.e., irregular and asymmetrical PSFs.\\
\indent The experiment settings are same to those in Subsection~\ref{subsec:rere}, except following three points: (1) all test images were blurred by two camera-shaking PSFs; (2) the first 36 order orthogonal functions were used; and (3) the facial images reconstructed by $\alpha$'s were also added into the gallery of candidate results.\\
\indent BRISQUE which is widely used to evaluate the quality of natural images which are assumed to be Gaussian distributed. However, as mentioned in Section~\ref{sec:intro}, the face images contain less strong edges, which means they may be not Gaussian distributed. The BRISQUE-L feature was reported to be relatively unaffected by small departures from model assumptions. Hence, we use BRISQUE-L feature for our task.\\
\indent The restoration results on CMU-PIE and extended Yale B datasets are shown in Fig.~\ref{fig:deblur_camera_pie} and Fig.~\ref{fig:deblur_camera_yale}, respectively. Our method significantly outperforms that of Krishnan~\emph{et~al.} In terms of restoration, our method has a tiny flaw on restoring shadow areas (the bottom two images in Fig.~\ref{fig:deblur_camera_yale}). This phenomenon is caused by collecting reconstructed images as candidate results. In reality, the trained SVR tends to give higher scores to images with less illumination variance. For example, the red numbers in Fig.~\ref{fig:deblur_camera_yale} are the scores of the original sharp images, while the blue numbers are the scores of the restoration results (in this case, the reconstructed images). The recognition results on both datasets are given in Table~\ref{tb:rec_camera}, which demonstrates the superiority of our method in the recognition of blurred facial images.\\

\begin{table}[h]
\begin{center}
\caption{Recognition rates}
\label{tb:rec_camera}
\begin{tabular}{|c|c|c|c|c|}
\hline
& \multicolumn{2}{|c|}{PIE} & \multicolumn{2}{|c|}{Yale B} \\
\hline
PSF & TYPE I & TYPE II & TYPE I & TYPE II\\
\hline
JRR~\cite{JRR} &  90.0 & 85.1 & 82.8 & 77.4 \\
\hline
Ours & 94.9 & 91.9 & 90.3 & 83.5\\
\hline
\end{tabular}
\end{center}
\end{table}
\setlength{\tabcolsep}{1.4pt}
\subsection{Experiments on real blur}
\begin{figure*}[t]
\centering
\includegraphics[width=0.8\linewidth]{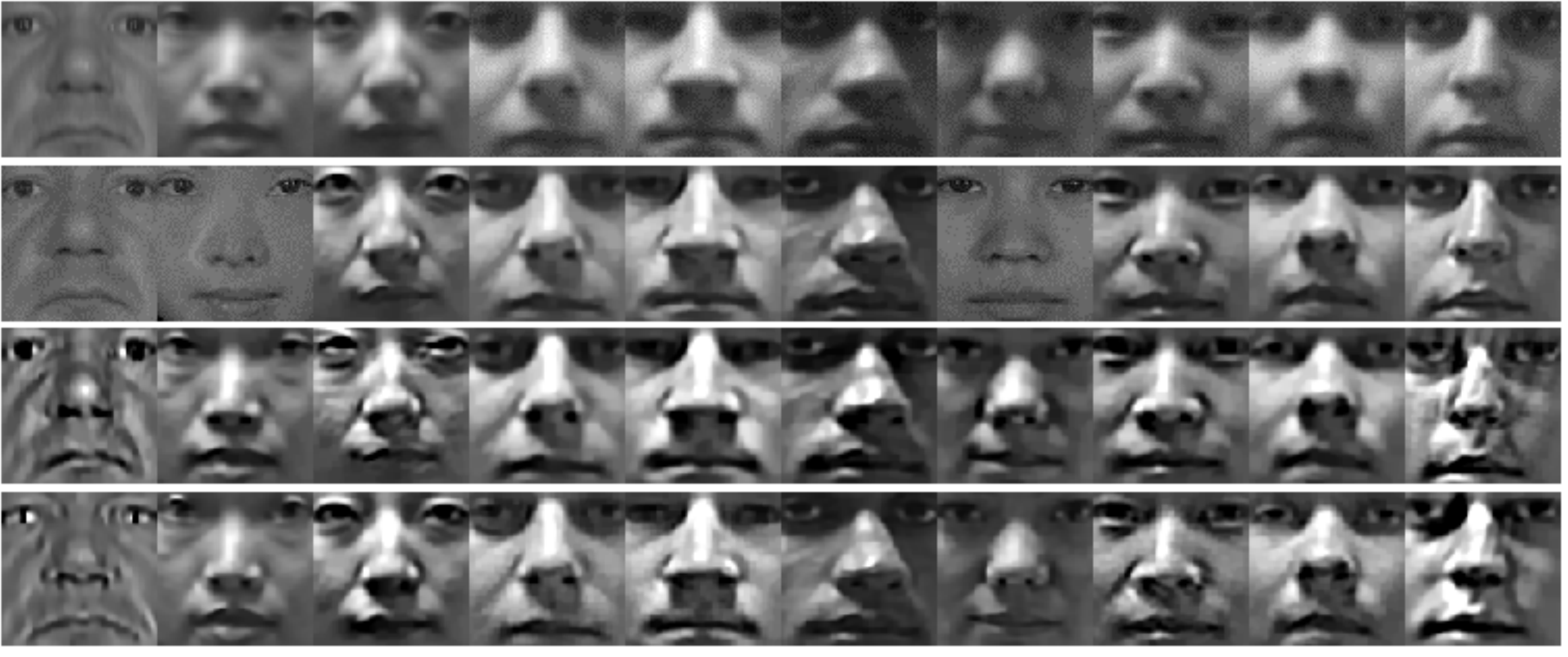}
\caption{The restoration results on FRGC 2.0. From top to bottom, they are OBs, and results of our method, Krishnan~\emph{et~al.}~\cite{Krishnan_CVPR2011} and Pan~\emph{et~al}~\cite{eccvguanshui2014}, respectively. The results marked by red rectangles are generated by the linear combinations of the function bases.}
\label{fig:real}
\end{figure*}

In this experiment, the restoration results of our method are compared with Krishnan \emph{et~al.} and Pan \emph{et~al}. and its recognition results are compared with JRR on FRGC 2.0 dataset. Again, FADEIN is not available for this case, because the PSF of real blur cannot be pre-defined in the training procedure.\\
\indent All the sharp images of each identity and the first 36 order orthogonal functions were used for constructing ${\bf A}$. The candidate results were generated in a similar way to that in Subsection~\ref{subsec:csb}, so did the training of SVR.\\
\indent The restoration results are shown in Fig.~\ref{fig:real}. In Fig.~\ref{fig:real}, our algorithm gave three linear combinations of face bases as the final restoration results which were marked by red rectangles. As there are no groudtruth sharp images for these OBs, the SSIM index is not available here. However, it is easy to visually observe the superiority of our restoration results against compared ones'. The recognition rates of JRR and our method are 93.5\% and 98.7\%, respectively.
\subsection{Computational efficiency}
Our algorithm gives the deblurring and recognition results at the same time. There are four major parts that take relative longer time to compute: 1) SVD; 2) construct A; 3) train SVR and 4) minimize Eq.~\eqref{eq:abfinal}. For each dataset, 1), 2) and 3) are just computed once. The computation complexity of SVD of an $P_1\times P_2$ matrix is $O(4P_1^2P_2+8P_1P_2^2+9P_2^3)$. In this case, $P_1$ is the number of image pixels and $P_2=MN$ where $M$ and $N$ are the number of face and function bases, respectively. To construct ${\bf A}$, $MN$ 2-D convolutions are needed. By using Fast Fourier Transformation (FFT), the computation complexity of 2-D convolution is $O(P_1P_2\log(P_1P_2))$. In this case, $P_1$ and $P_2$ are the height and width of an image. 3) and 4) are two constrained optimization problem which are solved iteratively. The convergence rates are highly dependent on the datasets and parameter settings. For all our experiments which were done on MATLAB 2014a on a PC with Intel Core i5 3.2GHz and 8GB RAM, 3) takes less than 1 second and 4) takes 20 to 40 seconds for one test image.

\section{Conclusions}
\label{sec:con}
In this paper, we proposed a coupled learning method combined with blind image quality assessment (BIQA) for image deconvolution. The method is specifically designed for deblurring face images that have few strong edges, and can theoretically estimate any PSF due to the reasonable assumptions and the adopted priors. We illustrate how to reduce computational costs for three kinds of symmetric PSF that are common in real applications. To illustrate how subsequent recognition tasks can be improved, we propose a new method that simultaneously generates deconvolution and recognition results. Experimentally, our proposed deconvolution method is superior to representative methods and the recognition method produces high recognition rates for blurred face images. In future work, we will focus on extending our PSF estimation strategy to natural image deblurring and our recognition method to non-frontal face recognition problems~\cite{ding2015comprehensive}.

\bibliographystyle{IEEEtran}
\bibliography{egbib}

\begin{thebibliography}{10}
\providecommand{\url}[1]{#1}
\csname url@samestyle\endcsname
\providecommand{\newblock}{\relax}
\providecommand{\bibinfo}[2]{#2}
\providecommand{\BIBentrySTDinterwordspacing}{\spaceskip=0pt\relax}
\providecommand{\BIBentryALTinterwordstretchfactor}{4}
\providecommand{\BIBentryALTinterwordspacing}{\spaceskip=\fontdimen2\font plus
\BIBentryALTinterwordstretchfactor\fontdimen3\font minus
  \fontdimen4\font\relax}
\providecommand{\BIBforeignlanguage}[2]{{%
\expandafter\ifx\csname l@#1\endcsname\relax
\typeout{** WARNING: IEEEtran.bst: No hyphenation pattern has been}%
\typeout{** loaded for the language `#1'. Using the pattern for}%
\typeout{** the default language instead.}%
\else
\language=\csname l@#1\endcsname
\fi
#2}}
\providecommand{\BIBdecl}{\relax}
\BIBdecl

\bibitem{Shan_SIGGRAPH2008}
Q.~Shan, J.~Jia, and A.~Agarwala, ``High-quality motion deblurring from a
  single image,'' in \emph{SIGGRAPH ASIA}, 2008, pp. 73:1--73:10.

\bibitem{Fergus_SIGGRAPH2006}
R.~Fergus, B.~Singh, A.~Hertzmann, S.~T. Roweis, and W.~T. Freeman, ``Removing
  camera shake from a single photograph,'' in \emph{SIGGRAPH}, 2006, pp.
  787--794.

\bibitem{gs1}
F.~Xue and T.~Blu, ``A novel sure-based criterion for parametric psf
  estimation,'' \emph{Image Processing, IEEE Transactions on}, vol.~24, no.~2,
  pp. 595--607, Feb 2015.

\bibitem{gs2}
M.~Delbracio and G.~Sapiro, ``Removing camera shake via weighted fourier burst
  accumulation,'' \emph{Image Processing, IEEE Transactions on}, vol.~24,
  no.~11, pp. 3293--3307, Nov 2015.

\bibitem{gs3}
L.~Xiao, J.~Gregson, F.~Heide, and W.~Heidrich, ``Stochastic blind motion
  deblurring,'' \emph{Image Processing, IEEE Transactions on}, vol.~24, no.~10,
  pp. 3071--3085, Oct 2015.

\bibitem{gs4}
H.~Liu, X.~Sun, L.~Fang, and F.~Wu, ``Deblurring saturated night image with
  function-form kernel,'' \emph{Image Processing, IEEE Transactions on},
  vol.~24, no.~11, pp. 4637--4650, Nov 2015.

\bibitem{Levin_PAMI2011}
A.~Levin, Y.~Weiss, F.~Durand, and W.~T. Freeman, ``Understanding blind
  deconvolution algorithms,'' \emph{Pattern Analysis and Machine Intelligence,
  IEEE Transactions on}, vol.~33, no.~12, pp. 2354--2357, 2011.

\bibitem{JRR}
H.~Zhang, J.~Yang, Y.~Zhang, and N.~M.~N. adn Thomas S.~Huang, ``Close the
  loop: Joint blind image restoration and recognition with sparse
  representation prior,'' in \emph{Proceedings of International Conference on
  Computer Vision}, 2011, pp. 770--777.

\bibitem{Liao_TIP2005}
Y.~Liao and X.~Lin, ``Blind image restoration with eigen-face subspace,''
  \emph{Image Processing, IEEE Transactions on}, vol.~14, no.~11, pp.
  1766--1772, 2005.

\bibitem{FADEIN}
M.~Nishiyama, A.~Hadid, H.~Takeshima, J.~Shotton, T.~Kozakaya, and
  O.~Yamaguchi, ``Facial deblur inference using subspace analysis for
  recognition of blurred faces,'' \emph{Pattern Analysis and Machine
  Intelligence, IEEE Transactions on}, vol.~33, no.~4, pp. 838--845, 2011.

\bibitem{jiangyou1}
G.~Tzimiropoulos, S.~Zafeiriou, and M.~Pantic, ``Sparse representations of
  image gradient orientations for visual recognition and tracking,'' in
  \emph{Proceedings of IEEE Computer Vision and Pattern Recognition, Workshop
  on CVPR for Human Behaviour Analysis}, 2011, pp. 26--33.

\bibitem{SparseRecognition1}
A.~Yang, Z.~Zhou, A.~Balasubramanian, S.~Sastry, and Y.~Ma, ``Fast $l_1$
  minimization algorithms for robust face recognition,'' \emph{Image
  Processing, IEEE Transactions on}, vol.~22, no.~8, pp. 3234--3246, Aug 2013.

\bibitem{SparseRecognition2}
J.~Wright, A.~Yang, A.~Ganesh, S.~Sastry, and Y.~Ma, ``Robust face recognition
  via sparse representation,'' \emph{Pattern Analysis and Machine Intelligence,
  IEEE Transactions on}, vol.~31, no.~2, pp. 210--227, Feb 2009.

\bibitem{SparseRecognition3}
X.~Zhao, X.~Chai, Z.~Niu, H.~C. Keng, and S.~Shan, ``Sparsely encoded local
  descriptor for face recognition,'' in \emph{International Conference on
  Automatic Face and Gesture Recognition}, 2011, pp. 149--154.

\bibitem{SparseDeblur2}
W.~Dong, D.~Zhang, G.~Shi, and X.~Wu, ``Image deblurring and super-resolution
  by adaptive sparse domain selection and adaptive regularization,''
  \emph{Image Processing, IEEE Transactions on}, vol.~20, no.~7, pp.
  1838--1857, Jul 2011.

\bibitem{SparseDeblur1}
W.~Dong, L.~Zhang, G.~Shi, and X.~Li, ``Nonlocally centralized sparse
  representation for image restoration,'' \emph{Image Processing, IEEE
  Transactions on}, vol.~22, no.~4, pp. 1620--1630, Apr 2013.

\bibitem{SparseDeblur3}
J.~Mairal, M.~Elad, and G.~Sapiro, ``Sparse representation for color image
  restoration,'' \emph{Image Processing, IEEE Transactions on}, vol.~17, no.~1,
  pp. 53--69, Jan 2008.

\bibitem{SparseDeblur4}
R.~Wang and D.~Tao, ``Recent progress in image deblurring,''
  \emph{arXiv:1409.6838v1}, 2014.

\bibitem{eccvguanshui2014}
J.~Pan, Z.~Hu, Z.~Su, and M.-H. Yang, ``Deblurring face images with
  exemplars,'' in \emph{European Conference on Computer Vision}, 2014, pp.
  47--62.

\bibitem{Chan_TIP2011}
S.~Chan, R.~Khoshabeh, K.~Gibson, P.~Gill, and T.~Nguyen, ``An augmented
  lagrangian method for total variation video restoration,'' \emph{Image
  Processing, IEEE Transactions on}, vol.~20, no.~11, pp. 3097--3111, Nov 2011.

\bibitem{BIQA}
A.~Mittal, A.~Moorthy, and A.~Bovik, ``Making image quality assessment
  robust,'' in \emph{IEEE Conference Record of the Forty Sixth Asilomar
  Conference on Signals, Systems and Computers (ASILOMAR)}, 2012, pp.
  1718--1722.

\bibitem{ding2014multi}
C.~Ding, J.~Choi, D.~Tao, and L.~S. Davis, ``Multi-directional multi-level
  dual-cross patterns for robust face recognition,'' \emph{IEEE Trans. Pattern
  Anal. Mach. Intell., doi: 10.1109/TPAMI.2015.2462338}, 2015.

\bibitem{conjgrad}
M.~R. Hestenes and E.~Stiefel, ``Methods of conjugate gradients for solving
  linear systems,'' \emph{Journal of Research of National Bureau of Standards},
  vol.~49, no.~6, pp. 409--436, 1952.

\bibitem{ALag}
M.~R. Hestenes, ``Multiplier and gradient methods,'' \emph{Journal of
  Optimization Theory and Applications}, vol.~4, pp. 303--320, 1969.

\bibitem{BRISQUE}
A.~Mittal, A.~Moorthy, and A.~Bovik, ``No-reference image quality assessment in
  the spatial domain,'' \emph{Image Processing, IEEE Transactions on}, vol.~21,
  no.~12, pp. 4695--4708, Dec 2012.

\bibitem{ImageFiltering}
A.~Bovik, T.~Huang, and J.~Munson, D.C., ``A generalization of median filtering
  using linear combinations of order statistics,'' \emph{Acoustics, Speech and
  Signal Processing, IEEE Transactions on}, vol.~31, no.~6, pp. 1342--1350, Dec
  1983.

\bibitem{MKL}
M.~Gonen and E.~Alpaydm, ``Multiple kernel learning algorithms,'' \emph{Journal
  of Machine Learning Research}, vol.~12, pp. 2211--2268, 2011.

\bibitem{SMO-MKL}
S.~Vishwanathan, Z.~Sun, N.~Theera-Ampornpunt, and M.~Varma, ``Multiple kernel
  learning and the {SMO} algorithm,'' in \emph{Advances in Neural Informatin
  Processing Systems}, Dec 2010, pp. 2361--2369.

\bibitem{smo1}
J.~C. Platt, ``Fast training of support vector machines using sequential
  minimal optimization,'' in \emph{Advances in Kernel Methods - Support Vector
  Learning}.\hskip 1em plus 0.5em minus 0.4em\relax MIT Press, Jan 1998.

\bibitem{smo2}
R.-E. Fan, P.-H. Chen, and C.-J. Lin, ``Working set selection using second
  order information for training support vector machines,'' \emph{Journal of
  Machine Learning Research}, vol.~6, pp. 1889--1918, 2005.

\bibitem{smo3}
S.~S. Keerthi, S.~K. Shevade, C.~Bhattacharyya, and K.~R.~K. Murthy,
  ``Improvements to platt's smo algorithm for svm classifier design,''
  \emph{Neural Computation}, vol.~13, no.~3, pp. 637--649, Mar 2001.

\bibitem{shapealginment}
Y.~Tsin and T.~Kannade, ``A correlation-based approach to robust point set
  registration,'' in \emph{Proceedings of European Conference on Computer
  Vision}, 2004, pp. 558--569.

\bibitem{FERET}
P.~Phillips, H.~Moon, S.~Rizvi, and P.~Rauss, ``The feret evaluation
  methodology for face-recognition algorithms,'' \emph{Pattern Analysis and
  Machine Intelligence, IEEE Transactions on}, vol.~22, no.~10, pp. 1090--1104,
  Oct 2000.

\bibitem{CMU-PIE}
T.~Sim, S.~Baker, and M.~Bsat, ``The cmu pose, illumination, and expression
  database,'' \emph{Pattern Analysis and Machine Intelligence, IEEE
  Transactions on}, vol.~25, no.~12, pp. 1615--1618, Dec 2003.

\bibitem{eYaleB}
A.~Georghiades, P.~Belhumeur, and D.~Kriegman, ``From few to many: illumination
  cone models for face recognition under variable lighting and pose,''
  \emph{Pattern Analysis and Machine Intelligence, IEEE Transactions on},
  vol.~23, no.~6, pp. 643--660, Jun 2001.

\bibitem{Krishnan_CVPR2011}
D.~Krishnan, T.~Tay, and R.~Fergus, ``Blind deconvolution using a normalized
  sparsity measure,'' in \emph{Proceedings of IEEE Conference on Computer
  Vision and Pattern Recognition}, 2011, pp. 233--240.

\bibitem{uiqi_ssim}
Z.~Wang and A.~Bovik, ``A universal image quality index,'' \emph{Signal
  Processing Letters, IEEE}, vol.~9, no.~3, pp. 81--84, Mar 2002.

\bibitem{fourindexes}
Y.~A.~Y. AI-Najjar and D.~C. Soong, ``Comparison of image quality assessment:
  {PSNR}, {HVS}, {SSIM}, {UIQI},'' \emph{International Journal of Scientific \&
  Engineering Research}, vol.~3, no.~8, Aug 2012.

\bibitem{LPQ_ICPR}
T.~Ahonen, E.~Rahtu, V.~Ojansivu, and J.~Heikkila, ``Recognition of blurred
  faces using local phase quantization,'' in \emph{Proceedings of International
  Conference on Pattern Recognition}, 2008, pp. 1--4.

\bibitem{ding2015comprehensive}
C.~Ding and D.~Tao, ``A comprehensive survey on pose-invariant face
  recognition,'' \emph{ACM Trans. Intell. Syst. Technol.}, 2015.

\end{thebibliography}
\begin{IEEEbiography}[{\includegraphics[width=1in,height=1.25in,clip,keepaspectratio]{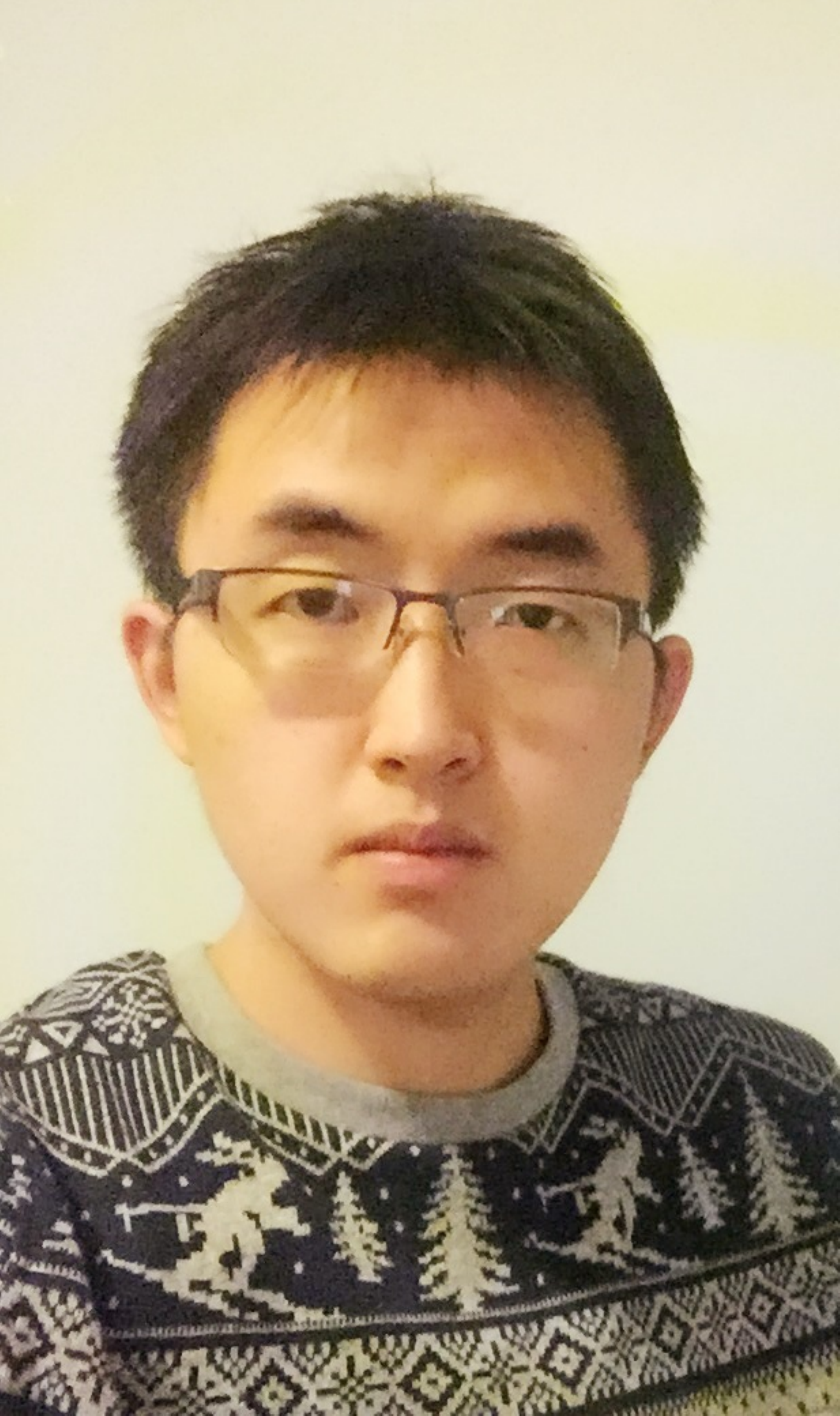}}]{Dayong Tian}
received the B.S. degree in Electronic Information Science and Technology and M.E. degree in Electronic Information Engineering from Xidian University, Xi'an, China. He is currently pursuing the Ph.D. degree with the Center for Quantum Computation and Intelligent Systems and the Faculty of Engineering and Information Technology, University of Technology at Sydney, Sydney, NSW, Australia. His research interests include computer vision and machine learning, and in particular, on image restoration, image retrieval and face recognition.
\end{IEEEbiography}
\begin{IEEEbiography}[{\includegraphics[width=1in,height=1.25in,clip,keepaspectratio]{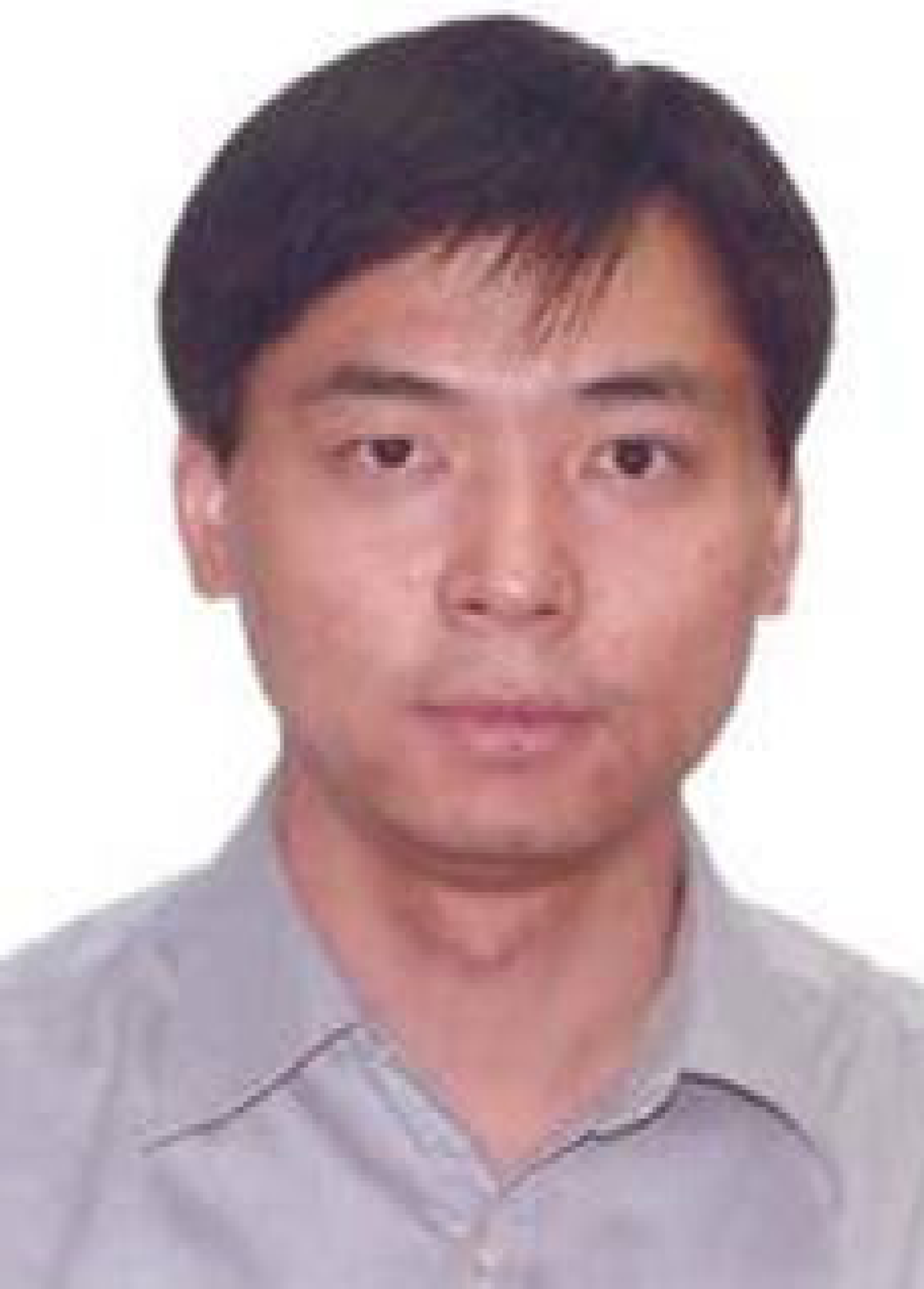}}]{Dacheng Tao}
(F' 15) is Professor of Computer Science with the Centre for Quantum Computation and Intelligent Systems, and the Faculty of Engineering and Information Technology in the University of Technology, Sydney. He mainly applies statistics and mathematics to data analytics and his research interests spread across computer vision, data science, image processing, machine learning, neural networks and video surveillance. His research results have expounded in one monograph and 100+ publications at prestigious journals and prominent conferences, such as IEEE T-PAMI, T-NNLS, T-IP, JMLR, IJCV, NIPS, ICML, CVPR, ICCV, ECCV, AISTATS, ICDM; and ACM SIGKDD, with several best paper awards, such as the best theory/algorithm paper runner up award in IEEE ICDM'07, the best student paper award in IEEE ICDM'13, and the 2014 ICDM 10 Year Highest Paper Award.
\end{IEEEbiography}

\end{document}